\documentclass[lettersize,journal]{IEEEtran}
\usepackage{amsmath,amsfonts}
\usepackage{algorithmic}
\usepackage{algorithm}
\usepackage{array}
\usepackage[caption=false,font=normalsize,labelfont=sf,textfont=sf]{subfig}
\usepackage{textcomp}
\usepackage{stfloats}
\usepackage{url}
\usepackage{verbatim}
\usepackage{graphicx}

\usepackage{amsthm}  
\usepackage{amsmath} 

\usepackage[normalem]{ulem}
\usepackage{multirow} 
\usepackage{pifont} 
\usepackage{hyperref}
\usepackage{booktabs} 

\usepackage[numbers]{natbib}

\newtheorem{definition}{Definition}
\newtheorem{theorem}{Theorem}

\newtheorem{corollary}[theorem]{Corollary}  

\usepackage{xr-hyper}

\newif\ifarxiv
\arxivtrue  

\usepackage{xr}
\usepackage{xcite}
\externalcitedocument{output}

\usepackage{amsmath,amsfonts,bm}

\def\eqref#1{equation~\ref{#1}}

\def\1{\bm{1}}

\def\rv{{\textnormal{v}}}

\def\mS{{\bm{S}}}
\def\mT{{\bm{T}}}

\def\mY{{\bm{Y}}}

\DeclareMathAlphabet{\mathsfit}{\encodingdefault}{\sfdefault}{m}{sl}
\SetMathAlphabet{\mathsfit}{bold}{\encodingdefault}{\sfdefault}{bx}{n}

\def\gA{{\mathcal{A}}}

\def\gE{{\mathcal{E}}}

\def\gR{{\mathcal{R}}}

\begin{document}

\title{CCE: Confidence-Consistency Evaluation for Time Series Anomaly Detection}

\author{Zhijie Zhong, Zhiwen Yu\IEEEauthorrefmark{1},~\IEEEmembership{Senior Member~IEEE}, Yiu-ming Cheung,~\IEEEmembership{Fellow IEEE}, Kaixiang Yang,~\IEEEmembership{Member~IEEE}
\IEEEcompsocitemizethanks{
\IEEEcompsocthanksitem Zhijie Zhong is with Pengcheng Laboratory, Shenzhen, Guangdong, 518066, China, and also with the School of Future Technology, South China University of Technology, Guangzhou, Guangdong 510650, China.
\IEEEcompsocthanksitem  Zhiwen~Yu is with the School of Computer Science and Engineering, South China University of Technology, Guangzhou, Guangdong 510650, China, and also with the Pengcheng Laboratory, Shenzhen, Guangdong 518066, China. Email: zhwyu@scut.edu.cn. Telephone number: 86-20-62893506. Fax number: 86-20-39380288.
\IEEEcompsocthanksitem Yiu-ming Cheung is with Department of Computer Science, Hong Kong Baptist University, Hong Kong SAR, China. E-mail: ymc@comp.hkbu.edu.hk
\IEEEcompsocthanksitem Kaixiang Yang is with the School of Computer Science and Engineering, South China University of Technology, Guangzhou, Guangdong 510650, China.
\IEEEcompsocthanksitem \IEEEauthorrefmark{1}Corresponding author: Zhiwen Yu.
}}


\markboth{Journal of \LaTeX\ Class Files,~Vol.~14, No.~8, August~2021}%
{Zhong \MakeLowercase{\textit{et al.}}: CCS: Confidence-Consistency Evaluation for Time Series Anomaly Detection}


\maketitle

\begin{abstract}
   Time Series Anomaly Detection metrics serve as crucial tools for model evaluation. However, existing metrics suffer from several limitations: insufficient discriminative power, strong hyperparameter dependency, sensitivity to perturbations, and high computational overhead. This paper introduces Confidence-Consistency Evaluation (CCE), a novel evaluation metric that simultaneously measures prediction confidence and uncertainty consistency. By employing Bayesian estimation to quantify the uncertainty of anomaly scores, we construct both global and event-level confidence and consistency scores for model predictions, resulting in a concise CCE metric. Theoretically and experimentally, we demonstrate that CCE possesses strict boundedness, Lipschitz robustness against score perturbations, and linear time complexity \(\mathcal{O}(n)\).
   Furthermore, we establish RankEval, a benchmark for comparing the ranking capabilities of various metrics. RankEval represents the first standardized and reproducible evaluation pipeline that enables objective comparison of evaluation metrics. Both CCE and RankEval implementations are fully open-source
   \footnote{Project site: \url{https://emorzz1g.github.io/CCE/}}.
\end{abstract}


\begin{IEEEkeywords}
Time Series Anomaly Detection, Evaluation, Bayesian Estimation, Uncertainty Estimation.
\end{IEEEkeywords}

\section{Introduction}

Time Series Anomaly Detection (TSAD) has extensive applications in industrial monitoring, cybersecurity, financial fraud detection, and other domains, aiming to identify anomalous patterns in time series data \cite{tpami1,tpami2,tpami3,tpami4,simads,cpatchbls}. With the advancement of deep learning technologies, an increasing number of state-of-the-art methods have been proposed for TSAD \cite{patchad,TFAD,couta,ImDiffusion,tpami5,tpami6,M2N2}. However, the current development of TSAD has reached a bottleneck, primarily due to the lag in evaluation metrics development compared to model advancement \cite{CurrentTimeSeries2022,simad,Navigating_the_metric_maze}. In early TSAD evaluations, classical metrics such as Precision, F1-score, and AUC-ROC were commonly used, but they primarily focused on the point-level detection capability of models. However, in TSAD, interval (event) anomalies are often of greater concern, making the evaluation of event-level detection capability particularly important. To address this, an increasing number of studies have proposed interval evaluation metrics to assess event-level detection capability, including R-based F1, F1 with point adjustment (F1-PA), Reduced-F1, Affiliation F1 (Aff-F1), VUS-ROC, PATE, and Unbiased Aff-F1 (UAff-F1) scores \cite{RbasedF1,F1PAscore,reducedPAF1score,affiliationF1,VUSROC,PATE,simad}. However, based on our analysis and previous literature, we have identified several limitations in these metrics:

\begin{enumerate}
   \item \textbf{Low Discriminability}: Previous studies \cite{CurrentTimeSeries2022,simad} have indicated that F1-PA \cite{F1PAscore} exhibits significant flaws when evaluating time series anomaly detection models. When faced with random scores, it fails to accurately reflect the actual performance of models, creating an illusion of progress in TSAD. Aff-F1 often yields scores greater than 0.67 in most cases, creating a false impression of good model performance. 
   
   \item \textbf{Hyperparameter Dependency}: F1-PA, Aff-F1, and UAff-F1 rely on anomaly threshold selection during evaluation, resulting in slow evaluation speed and limiting their use for model selection. VUS-ROC and PATE do not require anomaly thresholds but need event buffer size settings, which affects the stability and reproducibility of evaluation results and are unsuitable for scenarios with varying anomaly event lengths.
   
   \item \textbf{Low Robustness}: Existing evaluation metrics are sensitive to minor perturbations in anomaly scores, leading to unstable evaluation results. F1, F1-PA, and UAff-F1 all depend on anomaly score thresholds for calculation, making them highly sensitive to minor changes in anomaly scores and resulting in unstable evaluation outcomes.
   
   \item \textbf{High Computational Cost}: VUS-ROC and PATE metrics have high computational costs, limiting their application and evaluation on large-scale datasets. Typically, excessive complexity is not conducive to rapid validation of new model development.
   
   \item \textbf{Accuracy-oriented}: Existing evaluation metrics primarily focus on model detection accuracy, considering only the relationship between detection results and labels. Except for threshold-independent metrics such as AUC-ROC and VUS-ROC, other metrics ignore the detection consistency issue of models, often assuming that model anomaly scores are accurate and reliable. However, in practical applications, model predictions may contain uncertainty, especially in scenarios with high data noise or complex anomaly patterns.
\end{enumerate}

\begin{figure}[ht]
\centering
\includegraphics[width=1\linewidth]{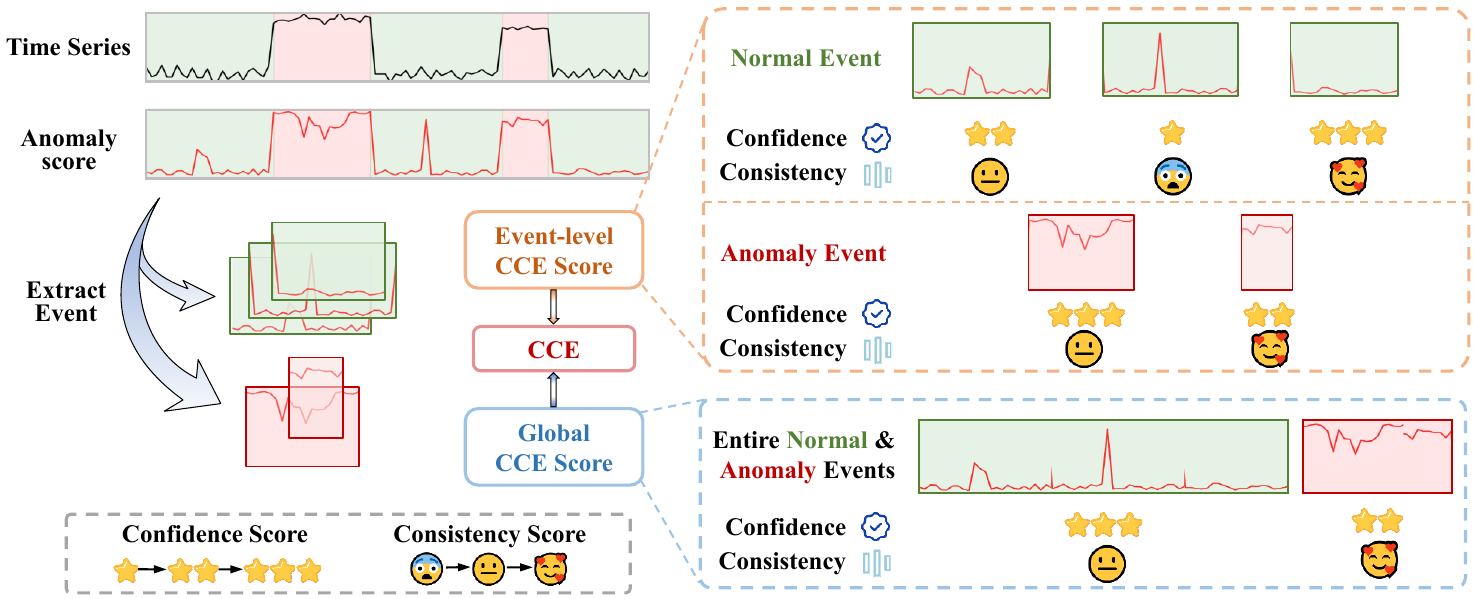}
\label{fig:fw}
\caption{Workflow of CCE framework. Note that the confidence and consistency score are continuous.}
\end{figure}

Furthermore, we found that current research lacks comprehensive and reproducible evaluation benchmarks for the metrics themselves, resulting in unclear limitations and applicability of different metrics. To address the above problems and metric limitations, we propose a novel evaluation metric called Confidence-Consistency Evaluation (CCE) and a benchmark for evaluating metrics, termed Rank-based evaluation benchmark (RankEval). To the best of our knowledge, CCE is the first metric that evaluates TSAD model performance by incorporating prediction confidence and uncertainty. We theoretically and experimentally demonstrate its robustness and computational efficiency. The fundamental motivation of CCE is that \textit{metrics should simultaneously consider prediction score confidence and consistency, rather than merely focusing on prediction results}. Specifically, we first model anomaly scores as Beta probability distributions and quantify the uncertainty of anomaly scores based on Bayesian statistical principles. We then convert uncertainty awareness into anomaly score consistency, and finally consider both event-level and global confidence and consistency to obtain the CCE score. The workflow of CCE is illustrated in Figure \ref{fig:fw}.

Moreover, to quantitatively and reproducibly evaluate the capabilities of different metrics, we designed the RankEval benchmark, motivated by a simple principle: \textit{the ideal model ranking should be consistent with the metric ranking}. The RankEval benchmark includes various time series anomaly detection datasets, encompassing both synthetic and real-world datasets, while evaluating the ranking capability, robustness, and computational efficiency of metrics.

We summarize our contributions as follows:
\begin{enumerate}
   \item We propose a novel confidence-consistency-based evaluation metric CCE for TSAD and a corresponding benchmark for evaluating metrics, termed RankEval, which contributes to further advancing the design of TSAD metrics and models.
   \item Through mathematical analysis, we verify numerous excellent properties of CCE, including boundedness, robustness, and low complexity.
   \item We validate the effectiveness of CCE on the RankEval benchmark and conduct comprehensive evaluation and analysis of multiple existing metrics. All our code is open-source\footnote{CCE's GitHub: \url{https://github.com/EmorZz1G/CCE}}, and we establish an automated standard evaluation process for updating the RankEval benchmark\footnote{RankEval: \url{https://emorzz1g.github.io/CCE/}}.
\end{enumerate}

\section{Related Work}
\subsection{Point-level Evaluation Metrics}
The most widely adopted point-level evaluation metrics are based on threshold-based binary classification approaches, with typical representatives including F1, precision, recall, and accuracy \cite{xblsg,HumanMoD,membls}. These metrics treat anomaly detection as a point-by-point binary classification problem, where the model first outputs an anomaly score sequence, then converts the scores to binary labels through thresholding, and finally performs point-by-point comparison with the ground truth labels. This approach is susceptible to noise and anomaly threshold selection. Considering the impact of anomaly threshold selection on evaluation accuracy, threshold-independent point-level metrics such as AUC-ROC and AUC-PR have also been proposed to evaluate the point-level detection capability of models. However, TSAD should focus more on the model's ability to detect entire intervals, and this evaluation approach ignores event-level assessment while overemphasizing point-level detection capability. Additionally, point-level metrics are extremely sensitive to annotation and anomaly score shifts, where slight temporal misalignment can cause dramatic fluctuations in metrics, thereby affecting evaluation reliability \cite{simad,CurrentTimeSeries2022,TSB-AD,TSB-AutoAD}.

\subsection{Event-level Evaluation Metrics}
To overcome the limitations of point-by-point metrics, previous research has treated continuous anomaly segments as events and proposed various event-level evaluation metrics. These metrics can be broadly categorized into three main approaches based on their underlying principles.

The first approach focuses on improving the original F1 score to directly evaluate interval anomalies. F1-PA \cite{F1PAscore} calculates F1 scores by performing point adjustment on detection results, thereby improving the interval evaluation capability of the original F1 metric. Building upon this concept, R-based F1 \cite{RbasedF1} evaluates the event-level detection capability of models by calculating the overlap rate between detection results and labels. Similarly, Reduced-F1 \cite{reducedPAF1score} computes F1 scores by reducing duplicate detected anomaly points, which mitigates the high-weight impact of long-interval anomalies.

The second approach emphasizes affiliation-based evaluation. Aff-F1 \cite{affiliationF1} evaluates the event-level detection capability of models by calculating the affiliation between detection results and anomaly events. To address data bias issues, UAff-F1 \cite{simad} improves upon Aff-F1 by eliminating such biases, thereby enhancing the discriminative capability of the original metric.

The third approach utilizes buffer-based evaluation strategies. VUS-ROC \cite{VUSROC} evaluates the event-level detection capability of models by adding event buffers and calculating the area under ROC curves at different thresholds. Following a similar principle, PATE \cite{PATE} also adds buffers and computes event-level precision and recall to assess model performance.

\subsection{Uncertainty Estimation}
Typically, the introduction of uncertainty helps models quantify the reliability of their predictions \cite{Bayesian2,Bayesian3,TFAD}. For example, AnoFormer \cite{uncertainty1} proposes using entropy to compute prediction uncertainty to clarify the decision boundary between normal and anomalous patterns. Further, TFAD \cite{TFAD} explores the principle of time-frequency domain uncertainty in temporal representations and proposes an anomaly detection method based on the time-frequency domain. LBAA \cite{uncertainty_landmark} and COUTA \cite{couta} both use calibrated anomalies to enhance the uncertainty estimation capability of models. ImDiffusion \cite{ImDiffusion} directly models anomaly uncertainty using diffusion models. Although uncertainty estimation has been validated in improving model detection capability, its application in the evaluation phase of time series anomaly detection remains unexplored.



\section{Methodology}
In this section, we first introduce the problem formulation and Bayesian uncertainty estimation method, then present the Confidence-Consistency Evaluation (CCE) Metric.

\subsection{Problem Formulation}
\begin{definition}[Problem Formulation]
Given a time series \( \mT = \{t_1, t_2, \ldots, t_n\} \) with corresponding anomaly scores \( \mS = \{s_1, s_2, \ldots, s_n\} \) and ground truth labels \( \mY = \{y_1, y_2, \ldots, y_n\} \) where \( y_i \in \{0, 1\} \), where \(n\) represents the number of time points, our goal is to evaluate the quality of anomaly detection models by quantifying both prediction confidence and uncertainty consistency. 
\end{definition}
It is worth noting that time series typically contain \(C\) channels, but our method is applicable to both univariate and multivariate time series anomaly detection, since CCE only evaluates anomaly scores.

\begin{definition}[Event Segmentation]
Let \( \mathcal{A} = \{\gA_1,\gA_2,\ldots,\gA_{N_a}\} \) denote the set of anomaly events and \( \mathcal{E} = \{\gE_1,\gE_2,\ldots,\gE_{N_e}\} \) denote the set of normal events, where \(\gA_i\) and \(\gE_j\) represent an anomaly event and normal event, respectively. For any event, we can use \(p\) and \(q\) to denote the start and end times. For brevity, we use \(\gE_i = (p, q)\) to represent an event, and \(|\gE_i|\) to represent the duration (length) of the event.
\end{definition}

\begin{theorem}[Uncertainty Estimation Function]
For a time series, such as anomaly scores \( \mS \), the uncertainty estimation function \( U: \mathbb{R}^n \rightarrow \mathbb{R} \) is defined as:
\begin{equation}
U(\mS) = \mathbb{E}[\text{Var}(\mS)]
\end{equation}
where \( \text{Var}(\mS) \) represents the variance of the prediction scores.
\end{theorem}

\subsection{Bayesian Uncertainty Estimation}
To construct the subsequent Confidence-Consistency Evaluation Framework, we need to perform uncertainty estimation on anomaly scores. For this purpose, we propose to use Bayesian uncertainty estimation \cite{Bayesian2,Bayesian3} as our primary method due to its theoretical soundness and practical advantages. The Bayesian approach provides a principled way to quantify uncertainty by modeling the underlying probability distribution of the prediction scores. We recommend the Bayesian approach for uncertainty estimation for three main reasons:

\begin{enumerate}
    \item \textbf{Natural Boundedness}: The Beta distribution naturally constrains values to [0,1], which aligns with normalized anomaly scores. In contrast, other distributions such as Gaussian distribution estimation have no value constraints, which cannot effectively constrain anomaly scores, ultimately leading to experimental errors due to different anomaly score scales.
    
    \item \textbf{Flexible Shape}: The Beta distribution can capture various shapes (symmetric, skewed, U-shaped) depending on the parameters, making it suitable for different types of anomaly patterns. In most cases, anomalies themselves are rare, which results in anomaly score distributions typically being skewed and dissimilar to Gaussian distributions.
    
    \item \textbf{Conjugate Properties}: The Beta distribution has desirable mathematical properties that facilitate uncertainty propagation and combination. If we use model output \(p_o\) to represent the probability of a certain anomaly, and our prior knowledge of anomaly scores is described by a Beta distribution, then after observing new data, the posterior distribution of \(p_o\) remains a Beta distribution.
\end{enumerate}

\subsubsection{Beta Distribution Modeling}

For a sequence of anomaly scores \( \mS = \{s_1, s_2, \ldots, s_n\} \) where \( s_i \in [0, 1] \), we model each event (a continuous segment of time points) as a Beta distribution parameterized by \( \alpha \) and \( \beta \). This approach treats each event as a collective distribution rather than modeling individual time points separately. The following statistical measures are defined:
(1) \( \overline{s} = \frac{1}{n}\sum_{i=1}^{n} s_i \) represents the mean of the anomaly scores.
(2) \( m_2 = \frac{1}{n}\sum_{i=1}^{n} (s_i - \overline{s})^2 \) represents the second central moment (variance) of the anomaly scores.

We model each sequence as a Beta distribution:
\begin{equation}
\mS \sim \text{Beta}(\alpha, \beta),
\end{equation}
where the parameters are defined using the method of moments:
\begin{equation}
\begin{aligned}
\alpha &= \overline{s}\left(\frac{\overline{s}(1-\overline{s})}{m_2}-1\right) \\
\beta &= (1-\overline{s})\left(\frac{\overline{s}(1-\overline{s})}{m_2}-1\right)
\end{aligned}
\label{eq:alpha_beta}
\end{equation}

\subsubsection{Uncertainty Estimation}
Given the Beta distribution parameters for each event, we can compute the uncertainty measure \(U\) for that event as:
\begin{equation}
U = \frac{\alpha \beta}{(\alpha + \beta)^2(\alpha + \beta + 1)}.
\end{equation}

This uncertainty measure represents the variance of the Beta distribution that models the specific event, providing an event-level measure of uncertainty for the anomaly detection model's outputs on that particular event segment.

\subsection{Confidence-Consistency Evaluation Framework}
The Confidence-Consistency Evaluation (CCE) framework is designed to assess the quality of anomaly detection models by evaluating both the confidence of predictions and the consistency of uncertainty estimates. The framework consists of three main components: event-level scoring, global scoring, and confidence-consistency evaluation computation.

\begin{definition}[Anomaly Event Confidence]
In TSAD, anomaly scores \(s_i\) are typically used to represent the probability that a time point \(t_i\) is anomalous. Therefore, for an anomaly event \(\gE_i = (p, q)\), we define the confidence of this anomaly event as:
\begin{equation}
   \text{Conf}(\gE_i) = \max(\frac{1}{|\gE_i|}\sum_{i=p}^{q} s_i  - \tau, 0).
\end{equation}
\end{definition}
Here, \(\tau\) represents the anomaly confidence threshold, with a default value of 0.5. The introduction of a confidence threshold is necessary because in practical applications, we typically only care about anomaly events with higher confidence levels. Furthermore, if we consider cases where the anomaly accuracy is below \(\tau\), the confidence of the anomaly event becomes:
\begin{equation}
   \text{Conf}^*(\gE_i) = \frac{1}{|\gE_i|}\sum_{i=p}^{q} s_i  - \tau.
\end{equation}
\label{def:anom_conf}
In the subsequent text, unless otherwise specified, we will refer to the confidence with an asterisk as relaxed confidence.

\begin{definition}[Normal Event Confidence]
Similarly, for normal events \( \gA_j = (p, q) \), we define the normal event confidence as:
\begin{equation}
   \text{Conf}(\gA_j) = \max(1 - \tau - \frac{1}{|\gA_j|}\sum_{j=p}^{q} s_j, 0).
\end{equation}
\label{def:norm_conf}
\end{definition}
Analogously, here \(1 - \tau\) represents the normal confidence threshold, meaning that the thresholds for normal and anomaly confidence are symmetric about 0.5. If we consider cases where the accuracy is below \(1-\tau\), the confidence of the normal event becomes:
\begin{equation}
   \text{Conf}^*(\gA_j) = 1 - \tau - \frac{1}{|\gA_j|}\sum_{j=p}^{q} s_j .
\end{equation}

\begin{definition}[Prediction Consistency]
For each event \( \gE_i \) or \( \gA_j \), we define the prediction consistency as:
\begin{equation}
   \begin{aligned}
   &\text{Cons}(\gE_i) = \exp(-\frac{1}{|\gE_i|}\sum_{k=p}^{q} U_k)\\
   &\text{Cons}(\gA_j) = \exp(-\frac{1}{|\gA_j|}\sum_{k=p}^{q} U_k).\\
   \end{aligned}
\end{equation}
\label{def:consistency}
\end{definition}
Where \( U_k \) is the uncertainty of the score at time \( k \). \(\text{Cons}(\gE_i)\) and \(\text{Cons}(\gA_j)\) represent the confidence consistency of anomaly events and normal events, respectively.
Intuitively, if an event has higher uncertainty, its consistency is lower, indicating that the model's prediction for that event is less reliable.

\subsubsection{Event-Level CCE Scoring}
To evaluate the assessment accuracy of the model for each anomaly event and normal event, we need to compute the CCE score for each event. For anomaly events \(\gE_i = (p, q)\) and normal events \(\gA_j = (p, q)\), we define their event-level scores as follows.

For each anomaly event \( \gE_i = (p, q) \), we compute the anomaly event score:
\begin{equation}
S_{\text{anom}}(\gE_i) = \text{Conf}(\gE_i) \times \text{Cons}(\gE_i)
\end{equation}
Similarly, for normal events \( \gA_j = (p, q) \), we compute the normal event score:
\begin{equation}
S_{\text{norm}}(\gA_j) = \text{Conf}(\gA_j) \times \text{Cons}(\gA_j)
\end{equation}

\begin{definition}[Event-Level Score]
The event-level score for anomaly and normal events, without loss of generality, is defined as:
\begin{equation}
S_{\text{event}} = \alpha \bar{S}_{\text{anom}} + (1-\alpha) \bar{S}_{\text{norm}}.
\label{def:event_score}
\end{equation}
\end{definition}
Where \( \bar{S}_{\text{anom}} \) and \( \bar{S}_{\text{norm}} \) are the average scores across all anomaly and normal events respectively, and \( \alpha \) is a weight parameter used to balance anomaly and normal events, typically set to 0.5, which can be adjusted based on detection requirements in practical scenarios.

\subsubsection{Global CCE Scoring}
To evaluate the model's performance across the entire time series, we design a global scoring mechanism. We treat all normal events across the entire time series as a comprehensive normal event, and all anomaly events across the entire time series as a comprehensive anomaly event. Specifically, \(\mS_{anom} = \{s_i \mid i \in \{\mathcal{A}\}\}\), where \(\{\mathcal{A}\}\) represents the set of time points for all anomaly events. Similarly, \(\mS_{norm} = \{s_j \mid j \in \{\mathcal{E}\}\}\), where \(\{\mathcal{E}\}\) represents the set of time points for all normal events.

\begin{definition}[Global Score]
The global CCE score for anomaly and normal events is defined as:
\begin{equation}
S_{\text{global}} = \eta S_{\text{anom}}^{\text{global}} + (1-\eta) S_{\text{norm}}^{\text{global}}
\end{equation}
where \( S_{\text{anom}}^{\text{global}} \) and \( S_{\text{norm}}^{\text{global}} \) are the global scores for anomaly and normal events respectively.
\label{def:global_score}
\end{definition}

Global scores for anomaly and normal events are computed as follows:
\begin{equation}
\begin{aligned}
&S_{\text{anom}}^{\text{global}} = \text{Conf}(\mS_{\text{anom}}) \times \text{Cons}(\mS_{\text{anom}})\\
&S_{\text{norm}}^{\text{global}} = \text{Conf}(\mS_{\text{norm}}) \times \text{Cons}(\mS_{\text{norm}})
\end{aligned}
\end{equation}

\begin{definition}[Confidence-Consistency Evaluation]
The final confidence-consistency evaluation score is defined as:
\begin{equation}
S_{\text{CCE}} = S_{\text{event}} + S_{\text{global}}
\end{equation}
\end{definition}
This score combines both event-level and global scores to provide a comprehensive evaluation of the model's performance. Finally, we present the pseudo-code implementation of CCE in Algorithm \ref{alg:cce}.

\begin{algorithm}[ht]
\caption{Confidence-Consistency Evaluation}
\label{alg:cce}
\begin{algorithmic}[1]
\REQUIRE Ground truth labels \( \mY \), anomaly scores \( \mS \), confidence thresholds \( \tau\), scale parameter \( \gamma \)
\ENSURE CCE score \( S_{\text{CCE}} \)
\STATE Normalize scores: \( \mS = \frac{\mS - \min(\mS)}{\max(\mS) - \min(\mS)} \)
\STATE Extract events: \( \mathcal{A}, \mathcal{E} = \text{extract\_events}(\mY) \)
\STATE Compute Beta parameters and uncertainties for each event using Eq. \ref{eq:alpha_beta}
\STATE Compute event-level CCE scores \( S_{\text{event}} \) using Definitions \ref{def:anom_conf}, \ref{def:norm_conf}, \ref{def:consistency}, and \ref{def:event_score}
\STATE Compute global CCE score \( S_{\text{global}} \) using Definitions \ref{def:global_score} and \ref{def:consistency}
\STATE Compute final CCE score: \( S_{\text{CCE}} = S_{\text{event}} + S_{\text{global}} \)
\RETURN \( S_{\text{CCE}} \)
\end{algorithmic}
\end{algorithm}

\section{Theory Analysis}\label{sec:theory}
In this section, we analyze the theoretical properties and computational complexity of CCE. CCE possesses the following properties, where Properties 1, 2, and 6 are straightforward to establish, while Properties 3, 4, and 5 are proven in this section.

\textbf{Property 1 (Intuition):} CCE quantifies the model's ability to distinguish between normal and anomalous events by evaluating both confidence and consistency. High confidence and high consistency indicate that the model's predictions for normal and anomalous events are more reliable.

\textbf{Property 2 (Interpretability):} CCE provides interpretable scores that can be used to identify model weaknesses and areas for improvement. By analyzing event-level and global scores, we can better understand the model's performance on different types of events, thereby guiding the direction of model improvement.

\textbf{Property 3 (Boundedness):} The final score \( S_{\text{CCE}} \in [0, 1] \), where 0 indicates poor performance and 1 indicates excellent performance. If we remove constraints to confidence, the CCE score has a lower bound of -1 and an upper bound of 1.

\textbf{Property 4 (Robustness):} CCE is robust to small perturbations in anomaly scores due to its Lipschitz continuity. This means that when anomaly scores undergo minor changes due to noise or computational errors, the change in the CCE metric will be strictly controlled within L times the magnitude of the perturbation.

\textbf{Property 5 (Scalability):} The computational complexity of CCE is \(\mathcal{O}(n)\), where \(n\) is the length of the time series data. CCE can be efficiently computed for large datasets, making it suitable for large-scale anomaly detection tasks.

\textbf{Property 6 (Symmetry):} The evaluation is symmetric with respect to anomaly and normal events, ensuring fair assessment. When \(\eta=0.5\), this property guarantees that the model's evaluation of anomaly and normal events is balanced, without bias toward any particular type of event.

\subsection{Proof of CCE Boundedness}
\begin{theorem}
The variance of a Beta distribution has an upper bound of \(\frac{1}{4}\).
\end{theorem}
\begin{proof}
We prove that the variance of a Beta distribution has an upper bound, with a maximum value of \(\frac{1}{4}\).

The probability density function of a Beta distribution depends on two shape parameters \(\alpha > 0\) and \(\beta > 0\), and its variance formula is:
\begin{equation}
   \text{Var}(\mS) = \frac{\alpha\beta}{(\alpha + \beta)^2 (\alpha + \beta + 1)},
\end{equation}
where \(\mS\) is a random variable following a Beta distribution, i.e., \(\mS \sim \text{Beta}(\alpha, \beta)\).

First, let \(n = \alpha + \beta\) and \(p = \frac{\alpha}{n}\) (where \(\beta = n(1-p)\) and \(0 < p < 1\)). The variance formula can be rewritten as:
\begin{equation}
   \text{Var}(\mS) = \frac{n p \cdot n(1-p)}{n^2 (n + 1)} = \frac{p(1-p)}{n + 1}
\end{equation}

For a fixed \(n\), the maximum value of \(p(1-p)\) is \(\frac{1}{4}\) (achieved when \(p = \frac{1}{2}\)). At this point, the variance simplifies to \(\frac{1}{4(n + 1)}\), which clearly decreases as \(n\) increases.

To find the global maximum, we need to consider the case when \(n\) approaches its minimum value. Since \(\alpha > 0\) and \(\beta > 0\), we have \(n > 0\). When \(n \to 0^+\) (i.e., both \(\alpha\) and \(\beta\) approach 0), the variance approaches \(\frac{1}{4}\).

Therefore, regardless of how \(\alpha\) and \(\beta\) are chosen (as long as they are positive), the variance of the Beta distribution will never exceed \(\frac{1}{4}\).
\end{proof}

\begin{corollary}
Both confidence and consistency scores are bounded.
\end{corollary}
\begin{proof}
For the event-level confidence of anomaly events, we have: \( \text{Conf}(\gA_i) = \max\left(\frac{1}{|\gA_i|}\sum_{k=p}^{q} s_k - \tau, 0\right) \in [0, 1-\tau] \). If we do not consider applying positive value constraints to confidence, then \(\text{Conf}^*(\gA_i) \in [-\tau, 1-\tau]\).

Similarly, for the event-level confidence of normal events, we have: \( \text{Conf}(\gE_j) = \max\left(1 - \frac{1}{|\gE_j|}\sum_{k=p}^{q} s_k - \tau, 0\right) \in [0, 1-\tau] \). If we do not consider applying positive value constraints to confidence, then \(\text{Conf}^*(\gE_i) \in [-\tau, 1-\tau]\).

Since the uncertainty \( U_k \in [0, 0.25] \) (the upper bound of Beta distribution variance is \( \frac{1}{4} \)), the event-level consistency satisfies:
\begin{equation}
   \begin{aligned}
      &\text{Cons}(\gA_i) = \exp\left(-\frac{1}{|\gA_i|}\sum_{k=p}^{q} U_k\right) \in [e^{-0.25}, 1]\\
      &\text{Cons}(\gE_j) \in [e^{-0.25}, 1]\\
   \end{aligned}
\end{equation}
Similarly, global confidence and consistency also satisfy the above bounds.
\end{proof}

\begin{theorem}
\(S_{\text{event}}\) and \(S_{\text{global}}\) are bounded.
\end{theorem}
\begin{proof}
If we consider applying positive value constraints to confidence, then:
\begin{equation}
   \begin{aligned}
   S_{\text{anom}}(\gA_i) = \text{Conf}^*(\gA_i) \times \text{Cons}(\gA_i) \in [0, 1-\tau]\\
   S_{\text{norm}}(\gE_j) = \text{Conf}^*(\gE_j) \times \text{Cons}(\gE_j) \in [0, 1-\tau]\\
   \end{aligned}
\end{equation}
If we consider relaxing the constraints on confidence, then:
\begin{equation}
   \begin{aligned}
   S^*_{\text{anom}}(\gA_i) = \text{Conf}(\gA_i) \times \text{Cons}(\gA_i) \in [-\tau, 1-\tau]\\
   S^*_{\text{norm}}(\gE_j) = \text{Conf}(\gE_j) \times \text{Cons}(\gE_j) \in [-\tau, 1-\tau]\\
   \end{aligned}
\end{equation}
Typically, we expect the model's confidence to be greater than 0.5, so we can assume \(\tau \in [0.5,1]\).
At this point, we have:
\begin{equation}
   S_{\text{anom}}(\gA_i), S_{\text{norm}}(\gE_j) \in [0, 0.5]
\end{equation}

For relaxed confidence \(S^*_{\text{anom}}\) and \(S^*_{\text{norm}}\), the maximum and minimum values are related to their corresponding confidence thresholds, but they always satisfy:
\begin{equation}
   S^*_{\text{anom}}(\gA_i), S^*_{\text{norm}}(\gE_j) \in [-0.5, 0.5]
\end{equation}

Since \(S_{\text{event}}\) and \(S_{\text{global}}\) are simple weighted functions, we have:
\begin{equation}
   \begin{aligned}
   &S_{\text{event}},S_{\text{global}} \in [0, 0.5]\\
   &S^*_{\text{event}},S^*_{\text{global}} \in [-0.5, 0.5]\\
   \end{aligned}
\end{equation}
\end{proof}

\begin{corollary}
CCE is bounded, with \(S_{\text{CCE}} \in [0, 1]\) and \(S_{\text{CCE}}^* \in [-1,1]\).
\end{corollary}
Since CCE is the sum of \(S_{\text{event}}\) and \(S_{\text{global}}\), this conclusion is straightforward to obtain.

\subsection{Proof of CCE Robustness}
Since the entire process of TSAD often involves a large amount of noise, where the most critical factor is the calculation error of anomaly scores, we need to prove that the CCE metric is stable against perturbations in anomaly scores. This ensures that when anomaly scores undergo minor changes due to noise or computational errors, the change in the CCE metric will be strictly controlled within \(L\) times the magnitude of the perturbation.

\begin{theorem}[CCE satisfies Lipschitz continuity against perturbations in anomaly scores]  
According to the assumption of Beta distribution modeling for anomaly scores, anomaly scores are normalized to \([0,1]\). Let \(\mathbf{s}=(s_1,\dots,s_n)\) and \(\tilde{\mathbf{s}}=(\tilde{s}_1,\dots,\tilde{s}_n)\) satisfy \(\|\mathbf{s}-\tilde{\mathbf{s}}\|_2\le\delta\).  
For \(S_{\mathrm{CCE}}\), there exists a constant \(L>0\), and \(L\) is close to 1, such that  
\[
|S_{\mathrm{CCE}}(\mathbf{s})-S_{\mathrm{CCE}}(\tilde{\mathbf{s}})|\le L\delta.
\]  
\end{theorem}

\begin{proof}
We prove that the CCE metric satisfies Lipschitz continuity against perturbations in anomaly scores.
Since CCE is composed of event-level scores and global scores, we need to prove that both event-level scores and global scores satisfy Lipschitz continuity against perturbations in anomaly scores.
For any anomaly event \(\mathcal E_i\) (interval \([p,q]\)), we explicitly calculate
\(
\bigl|\text{Conf}(\mathcal E_i)-\text{Conf}(\tilde{\mathcal E}_i)\bigr|
\)
and
\(
\bigl|\text{Cons}(\mathcal E_i)-\text{Cons}(\tilde{\mathcal E}_i)\bigr|
\)
respectively, then combine them to obtain the Lipschitz constant of event-level scores, and finally provide the global L.

First, for the difference in confidence between anomaly events:
\begin{equation}
\begin{aligned}
\bigl|\text{Conf}(\mathcal E_i)-\text{Conf}(\tilde{\mathcal E}_i)\bigr|
&\le \Bigl|\frac{1}{|\mathcal E_i|}\sum_{k=p}^{q}(s_k-\tilde s_k)\Bigr| \\
&\le\frac{1}{|\mathcal E_i|}\sum_{k=p}^{q}|s_k-\tilde s_k|
\le \delta.
\end{aligned}
\end{equation}
Similarly for normal events, the upper bound of the difference is still \(\delta\).

For each event \(\mathcal{E}_i\), the uncertainty \(U\) is computed from Beta distribution parameters. The sensitivity of uncertainty to individual score changes can be bounded by analyzing the partial derivatives with respect to the event-level statistics.

Now we compute the sensitivity of uncertainty to individual score changes. For any time point \(k \in [p,q]\), we have:
\begin{equation}
\begin{aligned}
\frac{\partial \overline{s}_{\mathcal{E}_i}}{\partial s_k} &= \frac{1}{|\mathcal{E}_i|} \\
\frac{\partial m_{2,\mathcal{E}_i}}{\partial s_k} &= \frac{2}{|\mathcal{E}_i|}(s_k - \overline{s}_{\mathcal{E}_i})
\end{aligned}
\end{equation}

Since \(|s_k - \overline{s}_{\mathcal{E}_i}| \leq 1\) (as \(s_k \in [0,1]\)), we have:
\begin{equation}
\left|\frac{\partial m_{2,\mathcal{E}_i}}{\partial s_k}\right| \leq \frac{2}{|\mathcal{E}_i|}
\end{equation}

Using the chain rule and the fact that \(U\) is a smooth function of \(\alpha\) and \(\beta\), and \(\alpha, \beta\) are smooth functions of \(\overline{s}_{\mathcal{E}_i}\) and \(m_{2,\mathcal{E}_i}\), we can bound the partial derivative of \(U\) with respect to \(s_k\):
\begin{equation}
\left|\frac{\partial U}{\partial s_k}\right| \leq \frac{C_1}{|\mathcal{E}_i|}
\end{equation}

where \(C_1\) is a constant that depends on the range of \(\alpha\) and \(\beta\) values. Since \(\alpha, \beta > 0\) and are bounded by the event size and score statistics, \(C_1\) is finite.

Therefore, for any event \(\mathcal{E}_i\), the change in uncertainty satisfies:
\begin{equation}
|U(\mathbf{s}_{\mathcal{E}_i}) - U(\tilde{\mathbf{s}}_{\mathcal{E}_i})| \leq \frac{C_1}{|\mathcal{E}_i|} \|\mathbf{s}_{\mathcal{E}_i} - \tilde{\mathbf{s}}_{\mathcal{E}_i}\|_2 \leq \frac{C_1}{|\mathcal{E}_i|} \delta
\end{equation}

Setting \(L_U = \frac{C_1}{|\mathcal{E}_i|}\), we have:
\begin{equation}
|U(\mathbf{s}_{\mathcal{E}_i}) - U(\tilde{\mathbf{s}}_{\mathcal{E}_i})| \leq L_U \delta
\end{equation}

To compute the exact value of \(C_1\), we analyze the partial derivatives more carefully. Let us define \(n = \alpha + \beta\) and \(p = \frac{\alpha}{n}\), then:
\begin{equation}
U = \frac{\alpha \beta}{n^2(n+1)} = \frac{p(1-p)}{n+1}
\end{equation}

The partial derivative with respect to \(s_k\) is:
\begin{equation}
\begin{aligned}
\frac{\partial U}{\partial s_k} &= \frac{\partial U}{\partial p} \cdot \frac{\partial p}{\partial s_k} + \frac{\partial U}{\partial n} \cdot \frac{\partial n}{\partial s_k} \\
&= \frac{1-2p}{n+1} \cdot \frac{\partial p}{\partial s_k} - \frac{p(1-p)}{(n+1)^2} \cdot \frac{\partial n}{\partial s_k}
\end{aligned}
\end{equation}

Since \(|p| \leq 1\) and \(|1-2p| \leq 1\), we have:
\begin{equation}
\left|\frac{\partial U}{\partial s_k}\right| \leq \frac{1}{n+1} \cdot \left|\frac{\partial p}{\partial s_k}\right| + \frac{1}{(n+1)^2} \cdot \left|\frac{\partial n}{\partial s_k}\right|
\end{equation}

From the definitions of \(\alpha\) and \(\beta\), we can show that:
\begin{equation}
\left|\frac{\partial p}{\partial s_k}\right| \leq \frac{2}{|\mathcal{E}_i|} \quad \text{and} \quad \left|\frac{\partial n}{\partial s_k}\right| \leq \frac{2}{|\mathcal{E}_i|}
\end{equation}

Therefore:
\begin{equation}
\left|\frac{\partial U}{\partial s_k}\right| \leq \frac{2}{|\mathcal{E}_i|(n+1)} + \frac{2}{|\mathcal{E}_i|(n+1)^2}
\end{equation}

To find a uniform upper bound that works for all possible values of \(n\), we need to analyze the function \(f(n) = \frac{2}{n+1} + \frac{2}{(n+1)^2}\) for \(n > 0\).

The derivative of \(f(n)\) with respect to \(n\) is:
\begin{equation}
f'(n) = -\frac{2}{(n+1)^2} - \frac{4}{(n+1)^3} < 0
\end{equation}

This shows that \(f(n)\) is a decreasing function of \(n\). The maximum value occurs at the minimum possible value of \(n\).

From the Beta distribution properties, we know that \(\alpha, \beta > 0\), which implies \(n = \alpha + \beta > 0\). However, for meaningful uncertainty estimation in practice, we typically have \(n \geq 1\). At \(n = 1\), we have:
\begin{equation}
f(1) = \frac{2}{2} + \frac{2}{4} = 1 + 0.5 = 1.5 \overset{\underset{\mathrm{def}}{}}{=} C_1.
\end{equation}


Thus,
\begin{equation}
|U_k-\tilde U_k|\le L_U|s_k-\tilde s_k|\le L_U\delta.
\end{equation}

Using the 1-Lipschitz property of \(\exp\) (on \([0,1]\)),
\begin{equation}
\begin{aligned}
\bigl|\text{Cons}(\mathcal E_i)-\text{Cons}(\tilde{\mathcal E}_i)\bigr|
&\le\Bigl|\frac{1}{\ell_i}\sum_{k=p}^{q}(U_k-\tilde U_k)\Bigr| \\
&\le\frac{1}{\ell_i}\sum_{k=p}^{q}L_U\delta=L_U\delta.
\end{aligned}
\end{equation}

Event-level scores are the product of two terms, i.e.,
\(
S_i=\text{Conf}(\mathcal E_i)\cdot\text{Cons}(\mathcal E_i).
\)
Since \(\text{Conf}(\mathcal{E}_i) \leq 1 - \tau\) and \(\text{Cons}(\mathcal{E}_i) \leq 1\), we use the product difference inequality
\(
|ab-\tilde a\tilde b|\le |a||b-\tilde b|+|\tilde b||a-\tilde a|,
\)
to obtain:
\[
|S_i-\tilde S_i|\le (1-\tau)\cdot L_U\delta+1\cdot\delta\le (0.5L_U+1)\delta.
\]

Subsequently, we can obtain the upper bound of the difference between normal and anomaly event-level scores:
\begin{equation}
   \begin{aligned}
      |\bar S_{\mathrm{norm}}-\tilde{\!\bar S}_{\mathrm{norm}}|\le (0.5L_U+1)\delta,\\
      |\bar S_{\mathrm{anom}}-\tilde{\!\bar S}_{\mathrm{anom}}|\le (0.5L_U+1)\delta.
\end{aligned}
\end{equation}

CCE scores are obtained by summing up event-level scores and global scores, i.e.,
\(
   S_{\text{CCE}} = \alpha \bar{S}_{\text{anom}} + (1 - \alpha) \bar{S}_{\text{norm}}
\)
Therefore, the perturbation amount of CCE scores is:
\begin{equation}
   \begin{aligned}
   &|S_{\text{CCE}} - \tilde{S}_{\text{CCE}}| \leq \alpha (0.5 L_U + 1) \delta + (1 - \alpha) (0.5 L_U + 1) \delta \\
   & = (0.5 L_U + 1) \delta.
\end{aligned}
\end{equation}
This shows that the CCE metric satisfies Lipschitz continuity against perturbations in anomaly scores, with the Lipschitz constant:
\(
L = 0.5 L_U + 1.
\)

Substituting \(L_U = \frac{C_1}{|\mathcal{E}_i|} = \frac{1.5}{|\mathcal{E}_i|}\), we get:
\begin{equation}
L = \frac{0.75}{|\mathcal{E}_i|} + 1
\end{equation}

For typical event sizes, we have:
\begin{itemize}
\item For events with length \(|\mathcal{E}_i| = 1\): \(L = \frac{0.75}{1} + 1 = 1.75\)
\item For events with length \(|\mathcal{E}_i| = 2\): \(L = \frac{0.75}{2} + 1 = 1.375\)
\item For events with length \(|\mathcal{E}_i| = 5\): \(L = \frac{0.75}{5} + 1 = 1.15\)
\item For events with length \(|\mathcal{E}_i| = 10\): \(L = \frac{0.75}{10} + 1 = 1.075\)
\item For events with length \(|\mathcal{E}_i| \geq 20\): \(L \leq 1.0375\)
\end{itemize}

As can be seen from Table \ref
{tab:metric_ana_config}, the 
average length of events in real-world scenarios 
is generally much greater than 20; therefore, \(L 
\leq 1.0375\). 

This ensures the stability of the CCE metric when there are errors in calculating anomaly scores. The Lipschitz constant is always close to 1, indicating that the CCE metric is robust to small perturbations in anomaly scores.

\end{proof}

\subsection{Computational Complexity}
To strictly analyze the computational complexity of CCE, we decompose it into the following six main steps, and provide the time complexity upper bound for each step.

\begin{theorem}[CCE Complexity]
Given a time series of length \(n\),
\(
   \mathcal{O}_{\text{CCE}} =  \mathcal{O}(n).
\)
\end{theorem}

\begin{proof}
We decompose the calculation of CCE into the following steps:

\begin{enumerate}
    \item \textbf{Event Extraction}:
    According to the given labels \( \mY \), we extract all anomaly events and normal events. This step can be completed through a single linear scan:
    \(
    \mathcal{O}(n)
    \)

    \item \textbf{Beta Parameter Computation}:
    For each anomaly score \( s_i \), we calculate its corresponding Beta distribution parameters \(\alpha_i\) and \(\beta_i\). Since the calculation of each score is independent, the complexity is:
    \(
    \mathcal{O}(n)
    \)

    \item \textbf{Uncertainty Estimation}:
    For each anomaly score, we calculate the variance \( U_i \) of its Beta distribution. Similarly, since each variance calculation is independent and requires only constant time:
    \(
    \mathcal{O}(n)
    \)

    \item \textbf{Event-Level Scoring}:
    For each event (anomaly or normal), we calculate its confidence and consistency scores. The complexity of each calculation is proportional to the length of the event, but the total length of all events is at most \( n \), therefore:
    \(
    \mathcal{O}(n)
    \)

    \item \textbf{Global Scoring}:
    We calculate the global confidence and consistency scores for the entire time series. This step only requires summing up all anomaly scores and normal scores respectively, therefore:
    \(
    \mathcal{O}(n)
    \)

    \item \textbf{CCE Score Computation}:
    We perform weighted summation of event-level scores and global scores, which requires only constant time:
    \(
    \mathcal{O}(1)
    \)
\end{enumerate}
In summary, all steps of CCE can be completed within \(\mathcal{O}(n)\) time, therefore:
\(
\mathcal{O}_{\text{CCE}} =  \mathcal{O}(n).
\)

\end{proof}











\section{Experiments}

In this section, we first introduce how to use the RankEval benchmark to evaluate different metrics. To provide a clear overview, the experiments are designed to address the following research questions (\textbf{RQs}):

\begin{itemize}
\item \textbf{RQ1}: How efficient are different metrics in terms of computational performance (Sec. \ref{sec:rq1_latency})?
\item \textbf{RQ2}: Does CCE have effectiveness and robustness compared to other metrics (Sec. \ref{sec:rq2_robust})?
\item \textbf{RQ3}: How do different metrics and models perform in terms of visualization (Sec. \ref{sec:rq3_vis})?
\item \textbf{RQ4}: Do the hyperparameters and task settings of CCE affect its evaluation capability (Sec. \ref{sec:rq4_hyper})?
\end{itemize}

\subsection{Experimental Setup}
In this section, we introduce the datasets, TSAD models, and evaluation methods used in RankEval.

\subsubsection{Datasets}
To ensure reproducibility of results, RankEval uses widely adopted and publicly available real-world datasets in the TSAD field, including MSL, SMD, PSM, SWAT, Creditcard, and UCR, among others \cite{simad,UCR}. Additionally, to verify the performance of different metrics under theoretical conditions, we constructed 30 synthetic datasets. The time series properties of these datasets are elaborated in Table \ref{tab:metric_ana_config} in Appendix.

 \subsubsection{TSAD Models}
 To analyze the performance of different metrics in real-world scenarios, we used two classic machine learning models (LOF \cite{LOF}, IForest \cite{IForest}) and five deep learning models (LSTMAD, USAD \cite{USAD}, AnomalyTransformer (A.T.) \cite{anom_trans}, TimesNet \cite{timesnet}, Donut). We used the public implementations of these models, and all models used default hyperparameter settings to ensure fairness and consistency of evaluation.
 
 Additionally, to analyze the performance of different metrics under theoretical conditions, we designed three different anomaly score generation models (ASGM) and their corresponding noise perturbation versions. For the default anomaly score generation models, we named them AccQ, LowDisAccQ, and PreQ-NegP. For the noise perturbation versions of anomaly score generation models (ASGM-R), we named them AccQ-R, LowDisAccQ-R, and PreQ-NegP-R. These ASGM-R models can be used to evaluate the robustness of different metrics. The specific ASGM-R models are as follows:
 
 \begin{enumerate}
    \item \textbf{AccQ}: The accuracy of this model is fixed at \(q\). If the true label is anomalous, then when correctly predicted (with probability \(q\)), the anomaly score is \(s^{(1)} = 0.9+0.1\mathcal{U}\); when incorrectly predicted (with probability \(1-q\)), the anomaly score is \(s^{(2)} = 0.05 \mathcal{U}\). If the true label is normal, then when correctly predicted, the anomaly score is \(s^{(1)}\); when incorrectly predicted, the anomaly score is \(s^{(2)}\). Here, \(\mathcal{U}\) represents the standard uniform distribution.
    
    \item \textbf{LowDisAccQ}: Similar to AccQ, this model's accuracy is fixed at \(q\), but the discriminative power of anomaly scores is lower: \(s^{(1)} = 0.6 + 0.1\mathcal{U}\), \(s^{(2)} = 0.4 \mathcal{U}\).
    
    \item \textbf{PreQ-NegP}: This model's anomaly precision is fixed at \(q\), and the false positive rate is fixed at \(p\). The model's default anomaly score is \(0.1\mathcal{U}\). For anomalies, with probability \(q\), the anomaly score is \(0.9\mathcal{U}+0.1\); for normal cases, with probability \(p\), the anomaly score is \(0.9\mathcal{U}+0.1\).
    
    \item \textbf{AccQ-R, LowDisAccQ-R, PreQ-NegP-R}: For the above three anomaly score generation methods, we add Gaussian noise with mean 0 and standard deviation \(\sigma\) to obtain the corresponding noise perturbation versions.
 \end{enumerate}
 
 Subsequently, when no ambiguity arises, we use the ASGM names to refer to the corresponding task types.
 
 \subsubsection{Evaluation Method}
 To evaluate the capability of different metrics in anomaly detection tasks, we adopt a ranking-based evaluation method, which is why our benchmark is called RankEval. This method can effectively measure the ranking capability of metrics for model performance, i.e., whether metrics can correctly rank models with better performance ahead of those with worse performance.
 
 Given an expected ranking \(\gR^* = [r_1^*, r_2^*, \ldots, r_n^*]\) and an actual ranking \(\gR = [r_1, r_2, \ldots, r_n]\) generated by some metric, we use the following three indicators to evaluate ranking quality:
 \begin{enumerate}
     \item \textbf{Spearman's Rank Correlation} \cite{spearman1987proof}:
     Measures the linear correlation between two rankings, defined as:
     \begin{equation}
         \text{Sp} = 1 - \frac{6 \sum_{i=1}^{n} (r_i^* - r_i)^2}{n(n^2 - 1)}
     \end{equation}
     where \(r_i^*\) and \(r_i\) respectively represent the ranking of the \(i\)-th model in the expected ranking and actual ranking. \(\rho_s \in [-1, 1]\), with values closer to 1 indicating more consistent rankings.
     \item \textbf{Kendall's Tau} \cite{kendall1938new}:
     Measures the agreement between two rankings, defined as:
     \begin{equation}
       \text{Kd} = \frac{C - D}{C + D}
     \end{equation}
     where \(C\) and \(D\) respectively represent the number of concordant pairs and discordant pairs. For any two model pairs \((i, j)\), if \(i\) appears before \(j\) in the expected ranking and also appears before \(j\) in the actual ranking, it is considered a concordant pair; otherwise, it is considered a discordant pair. \(\tau \in [-1, 1]\), with values closer to 1 indicating more consistent rankings.
     \item \textbf{Mean Rank Deviation} \cite{MDrank}:
     Measures the average deviation between the actual ranking and expected ranking, defined as:
     \begin{equation}
         \text{MD} = \frac{1}{n} \sum_{i=1}^{n} |r_i^* - r_i|
     \end{equation}
     where \(|r_i^* - r_i|\) represents the absolute difference in rankings of the \(i\)-th model between the two rankings. Smaller MD values indicate more accurate rankings.
 \end{enumerate}
 
 This evaluation method is more interpretable, with ranking results that are intuitive and easy to understand, facilitating the analysis of metric effectiveness. Moreover, it avoids the influence of differences in numerical ranges and distributions among different metrics.

\subsection{Computational Latency Analysis (\textbf{RQ1})}\label{sec:rq1_latency}

\subsubsection{Latency Performance}
\begin{figure}
    \centering
    \includegraphics[width=0.95\linewidth]{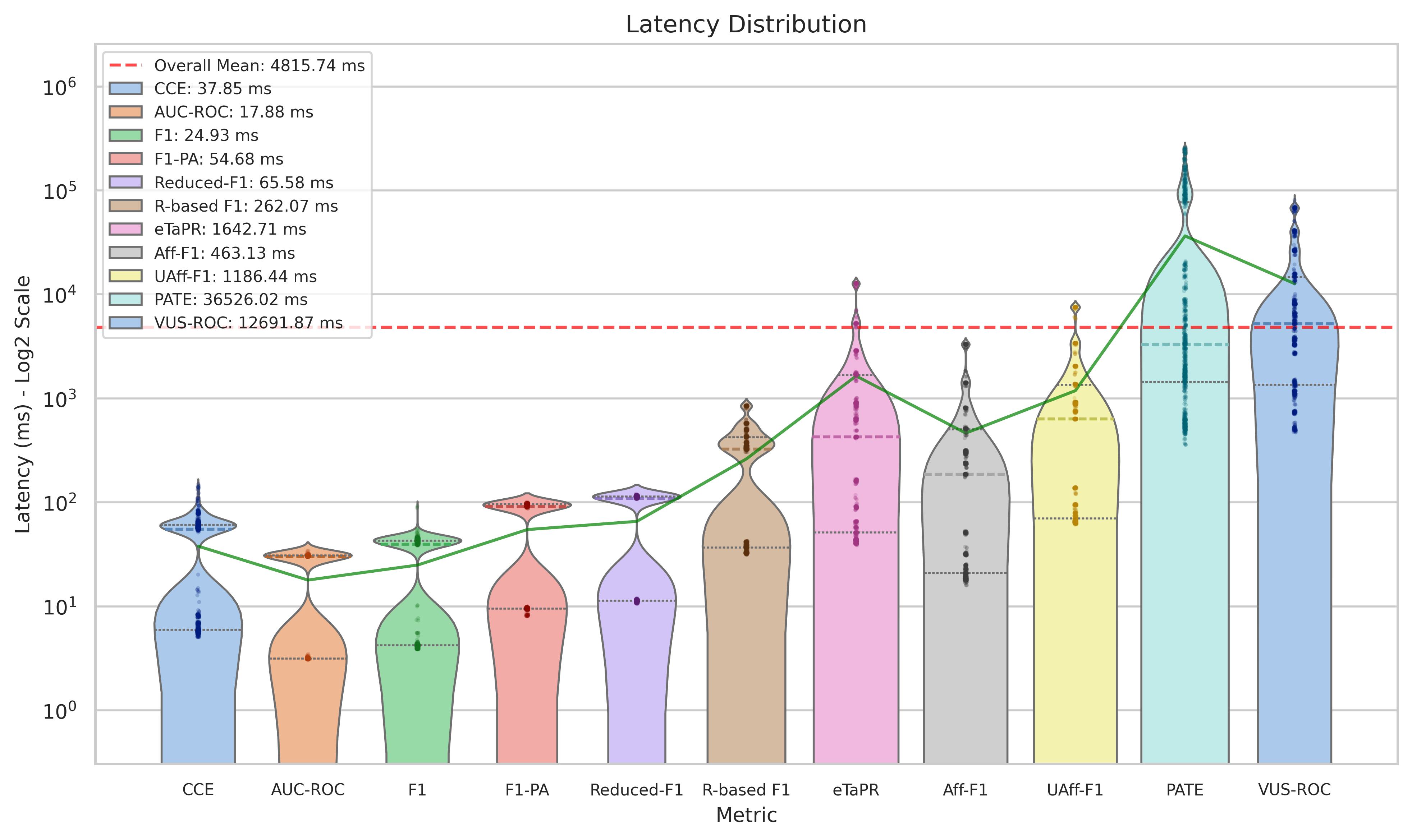}
    \caption{Latency distribution of different metrics. The violin plot shows the median and interquartile range. The green line represents the average latency of each metric.}
    \label{fig:latency}
\end{figure}

To analyze the computational efficiency of different metrics, we evaluated the latency distribution across all metrics on the AccQ task, as illustrated in Figure \ref{fig:latency}. Among all evaluated metrics, PATE exhibited the highest latency with an average of 36,526.02 ms, followed by VUS-ROC at 12,691.87 ms. In contrast, CCE demonstrated the lowest latency among interval-based metrics, achieving an average of only 37.85 ms. Remarkably, CCE's computational overhead is comparable to point-level metrics such as F1 and AUC-ROC. As a result, CCE provides significant benefits in computing as an interval metric, improving efficiency by up to \textbf{965 times}.

\subsubsection{CCE Computational Efficiency Analysis}
\begin{figure*}[htb]
   \subfloat[\label{fig:latency_task}]{
      \includegraphics[width=0.33\linewidth]{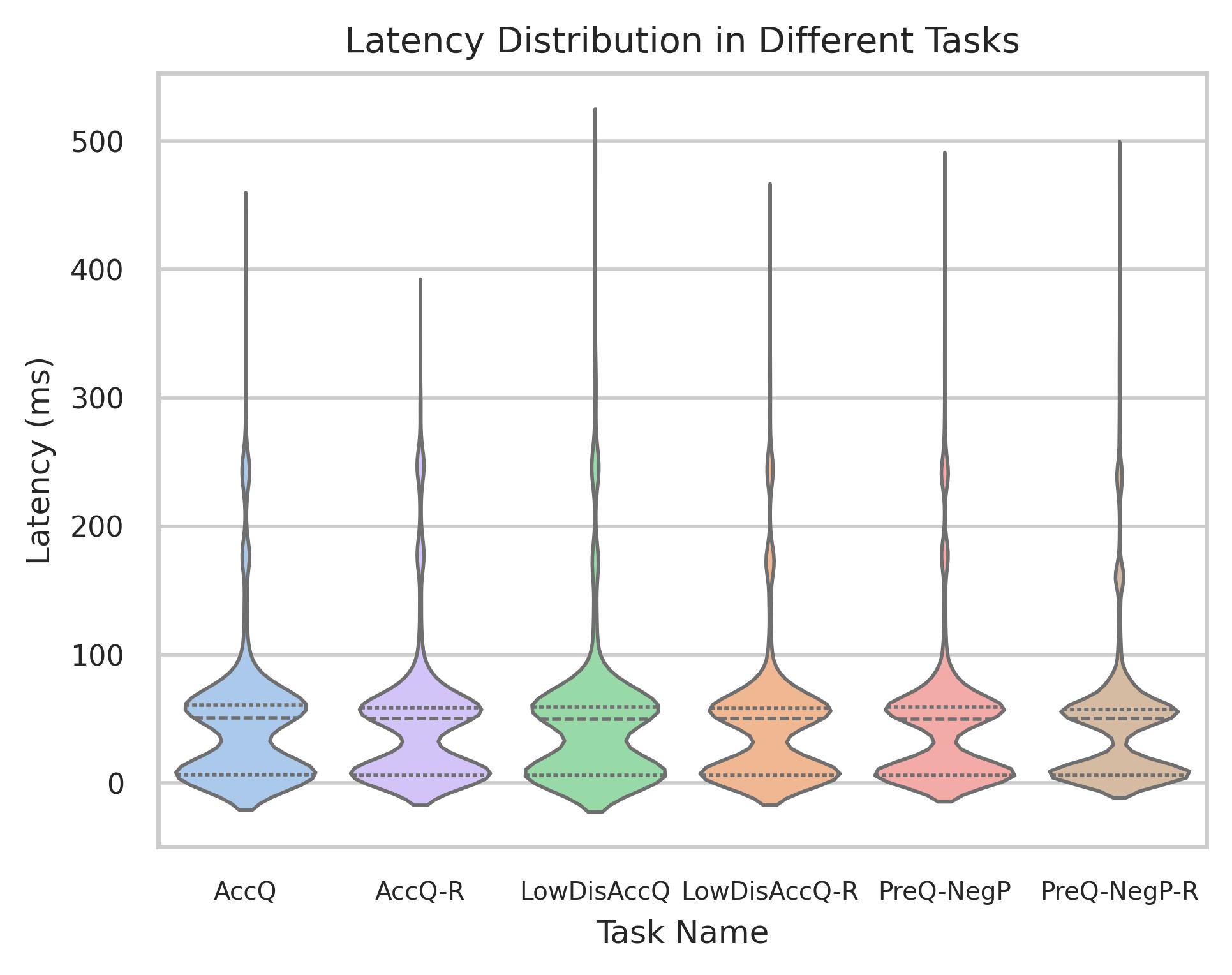}
   }
   \subfloat[\label{fig:latency_ts_len}]{
      \includegraphics[width=0.33\linewidth]{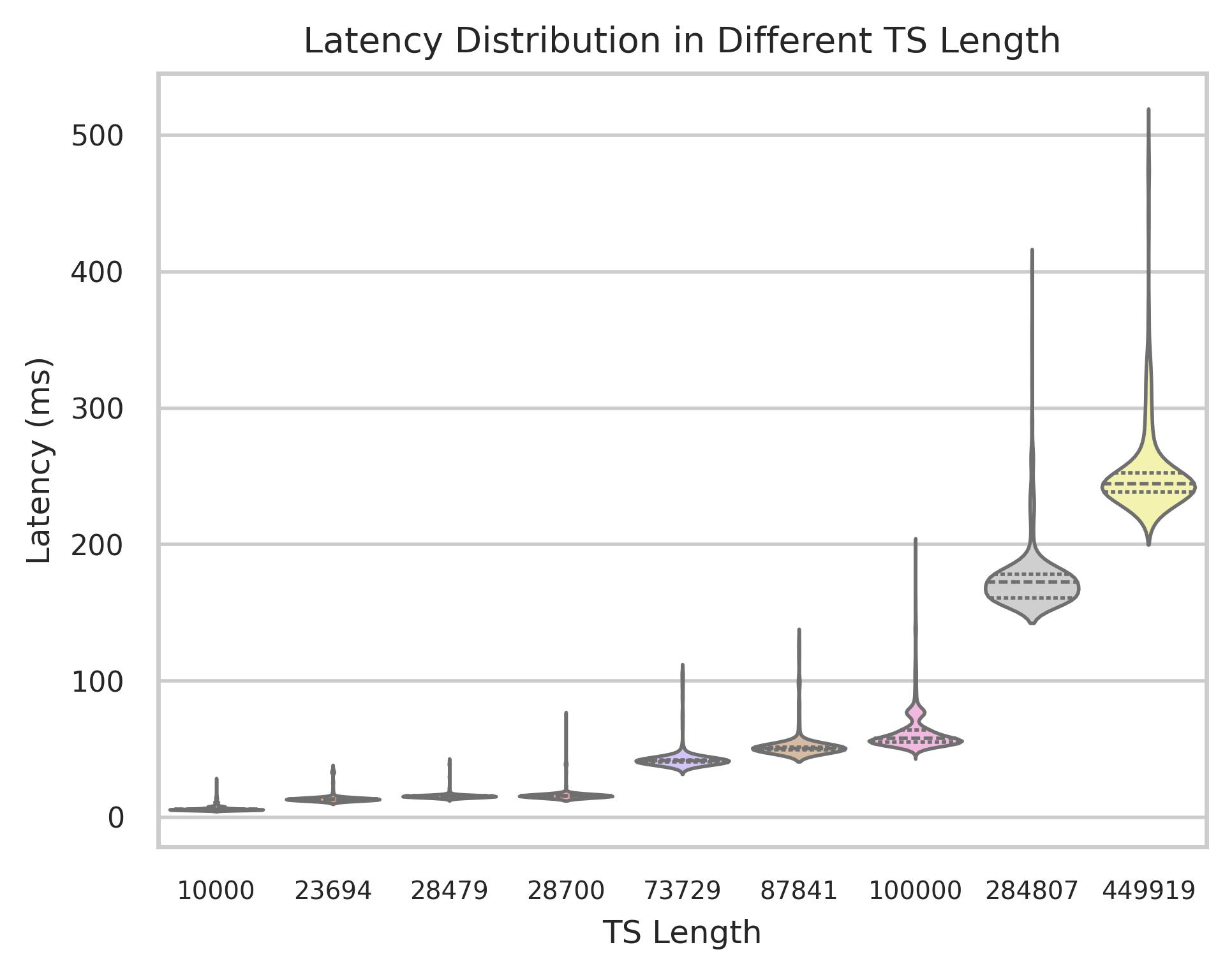}
   }
   \subfloat[\label{fig:latency_seg}]{
      \includegraphics[width=0.33\linewidth]{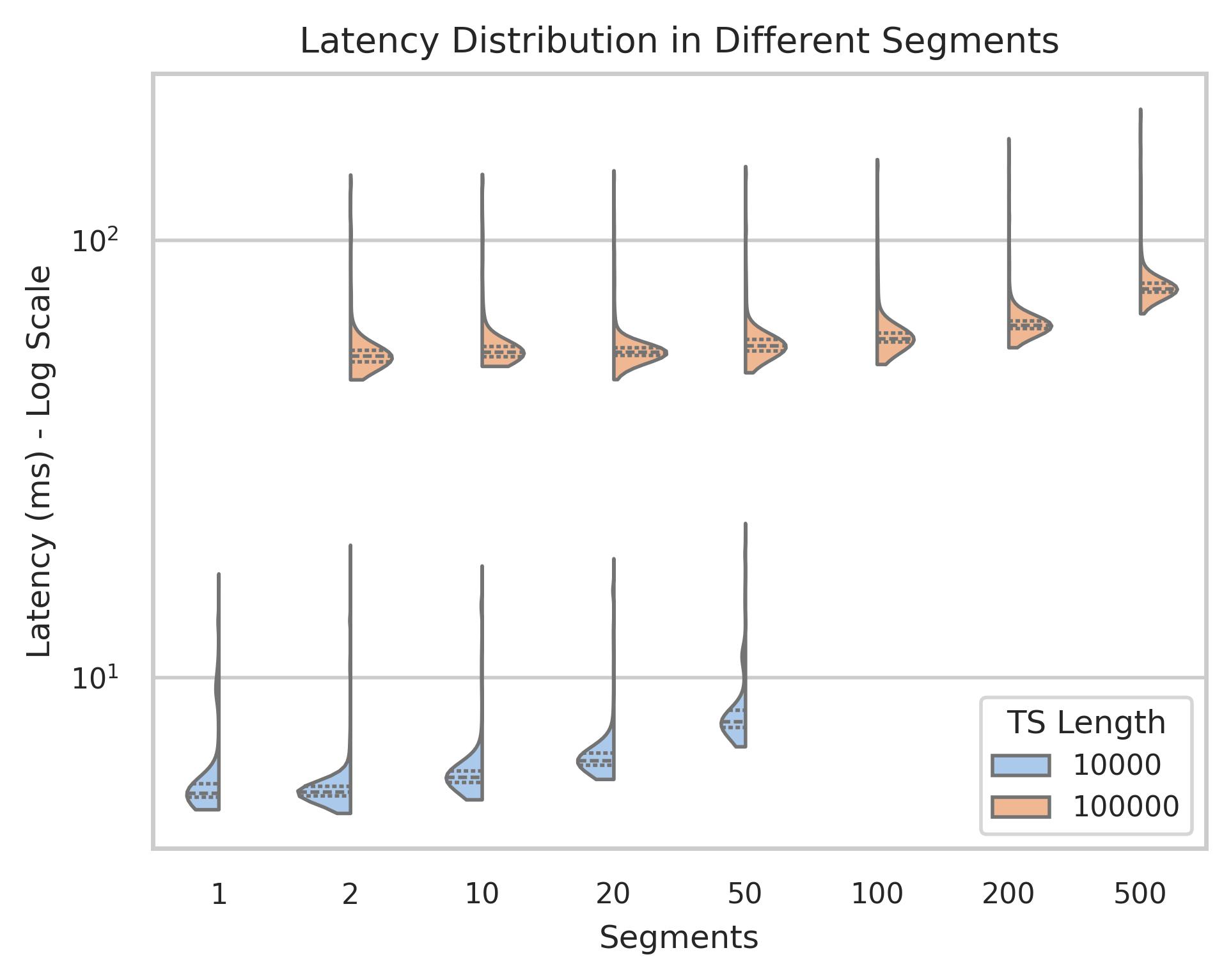}
   }
   \caption{CCE latency analysis under different scenarios: (a) task type impact, (b) time series length impact, and (c) anomaly segment count impact.}
\end{figure*}

Figures \ref{fig:latency_task}, \ref{fig:latency_ts_len}, and \ref{fig:latency_seg} respectively demonstrate the latency distribution of CCE under different scenarios. From Figure \ref{fig:latency_task}, it is clear that CCE's computational efficiency is independent of task type, with consistent latency distributions. Figure \ref{fig:latency_ts_len} shows the impact of different time series lengths on CCE latency. We find that CCE's latency increases approximately linearly, which is consistent with our previous theoretical analysis. Figure \ref{fig:latency_seg} demonstrates the impact of the number of anomalies on CCE latency. Overall, the number of event segments has little impact on CCE, which can be approximated as a constant, depending on the computational capabilities of the computer.

\subsection{Metric Effectiveness and Robustness (\textbf{RQ2})}\label{sec:rq2_robust}
To verify the effectiveness and robustness of CCE compared to other metrics, we analyzed the performance of different metrics under various parameter settings: \(p \in [0.1, 0.2, \ldots, 0.9, 1.0]\), \(q \in [0.01, 0.05, 0.1, 0.3]\), and \(\sigma \in [0.0, 0.05, 0.1]\).

\subsubsection{Effectiveness Verification}
For synthetic tasks, anomaly scores are generated by explicitly given parameters \((q, p)\) in ASGM.
Therefore, we can directly obtain the expected performance ranking of each model under ideal conditions based on \((q, p)\) and form the expected ranking \(\gR^*\). Specifically: AccQ(-R) and LowDisAccQ(-R) are solely determined by accuracy \(q\), with \(\gR^*\) arranged in ascending order by \(q\); PreQ-NegP(-R) gives rise to two single-factor tasks: PreQ-NegP-Q only examines anomaly precision \(q\), with \(\gR^*\) arranged in ascending order by \(q\); PreQ-NegP-P only examines false positive rate \(p\), with \(\gR^*\) arranged in descending order by \(p\) (lower false positive rate is better), ignoring the other dimension. Subsequently, we compare the rankings generated by each metric for the same set of models with \(\gR^*\), calculating Spearman, Kendall, and MD scores to measure metric capability.

The results are shown in Table \ref{tab:rank_comp}. Overall, R-based F1 has the worst ranking consistency capability, with an Sp score of 0.666. The ranking capabilities of F1, F1-PA, Reduced-F1, and eTaPR show small differences, but compared to more advanced metrics like Aff-F1, UAff-F1, and VUS-ROC, they still lag behind by more than 10\%, making these metrics somewhat outdated. Surprisingly, AUC-ROC is not only suitable for point anomalies but also performs excellently in interval anomaly scenarios. Ultimately, from a holistic perspective, CCE is the best metric across all scenarios, maintaining consistent ranking consistency. VUS-ROC and AUC-ROC are both recommended metrics, while (U)Aff-F1 is only recommended for use in specific scenarios (we will analyze the reasons in the next section).

It should be noted that AUC-ROC and VUS-ROC are not perfect metrics. Further analysis in Appendix \ref{app:noise_effect} (Figs. \ref{fig:robust_aucroc} and \ref{fig:robust_vusroc}) reveals that in noisy scenarios, false positives affect the ranking capability of these metrics, as they pay more attention to the model's anomaly prediction capability.

\begin{table*}[htbp]
   \centering
   \caption{Comparison of ranking consistency across different indicators (the larger the Sp and Kd, the better; the smaller the MD, the better). \textbf{bold} indicates the best, \uline{underlined} indicates the second best.}
   \resizebox{0.95\linewidth}{!}{
     \begin{tabular}{c|c|cccccccccc}
     \toprule
     \textbf{Task} & \textbf{Score} & \textbf{CCE} & \textbf{AUC-ROC} & \textbf{F1} & \textbf{F1-PA} & \textbf{Reduced-F1} & \textbf{R-based F1} & \textbf{eTaPR} & \textbf{Aff-F1} & \textbf{UAff-F1} & \textbf{VUS-ROC} \\
     \midrule
     AccQ  & Sp    & \textbf{1.000 } & \textbf{1.000 } & 0.340  & 0.340  & 0.340  & 0.294  & 0.780  & 0.832  & 0.825  & \textbf{1.000 } \\
     AccQ  & Kd    & \textbf{1.000 } & \textbf{1.000 } & 0.248  & 0.248  & 0.248  & 0.193  & 0.697  & 0.789  & 0.779  & \textbf{1.000 } \\
     AccQ  & MD    & \textbf{0.000 } & \textbf{0.000 } & 2.094  & 2.094  & 2.095  & 2.344  & 1.147  & 0.587  & 0.622  & \textbf{0.000 } \\
     \midrule
     LowDisAccQ & Sp    & \textbf{1.000 } & \textbf{1.000 } & 0.998  & 0.998  & 0.996  & 0.901  & 0.846  & 0.953  & 0.944  & \textbf{1.000 } \\
     LowDisAccQ & Kd    & \textbf{1.000 } & \textbf{1.000 } & 0.993  & 0.993  & 0.990  & 0.895  & 0.760  & 0.927  & 0.915  & \textbf{1.000 } \\
     LowDisAccQ & MD    & \textbf{0.000 } & \textbf{0.000 } & 0.025  & 0.025  & 0.038  & 0.465  & 0.933  & 0.265  & 0.299  & \textbf{0.000 } \\
     \midrule
     PreQ-NegP-Q & Sp    & \textbf{1.000 } & \textbf{1.000 } & 0.928  & 0.925  & 0.895  & 0.681  & 0.878  & 0.883  & 0.872  & \textbf{1.000 } \\
     PreQ-NegP-Q & Kd    & \textbf{1.000 } & \uline{0.999 } & 0.931  & 0.925  & 0.892  & 0.674  & 0.866  & 0.855  & 0.839  & \textbf{1.000 } \\
     PreQ-NegP-Q & MD    & \textbf{0.000 } & \uline{0.003 } & 0.267  & 0.287  & 0.371  & 1.131  & 0.494  & 0.551  & 0.619  & \uline{0.002 } \\
     \midrule
     PreQ-NegP-P & Sp    & \textbf{1.000 } & 0.987  & \textbf{1.000 } & \textbf{1.000 } & \textbf{1.000 } & 0.789  & 0.876  & 0.920  & 0.891  & 0.990  \\
     PreQ-NegP-P & Kd    & \textbf{1.000 } & 0.982  & \textbf{1.000 } & \textbf{1.000 } & \textbf{1.000 } & 0.776  & 0.864  & 0.895  & 0.863  & 0.986  \\
     PreQ-NegP-P & MD    & \textbf{0.000 } & 0.026  & \textbf{0.000 } & \textbf{0.000 } & \textbf{0.000 } & 0.267  & 0.160  & 0.150  & 0.192  & 0.019  \\
     \midrule
     Avg.  & Sp    & \textbf{1.000 } & 0.997  & 0.816  & 0.816  & 0.808  & 0.666  & 0.845  & 0.897  & 0.883  & \uline{0.998 } \\
     Avg.  & Kd    & \textbf{1.000 } & 0.995  & 0.793  & 0.792  & 0.783  & 0.635  & 0.797  & 0.866  & 0.849  & \uline{0.996 } \\
     Avg.  & MD    & \textbf{0.000 } & 0.007  & 0.597  & 0.601  & 0.626  & 1.052  & 0.684  & 0.388  & 0.433  & \uline{0.005 } \\
     \bottomrule
     \end{tabular}}
   \label{tab:rank_comp}%
\end{table*}%
 
\subsubsection{Robustness Analysis}
\begin{figure}[htb]
   \centering
   \subfloat[\label{fig:accq_rank}]{
   \includegraphics[width=1\linewidth]{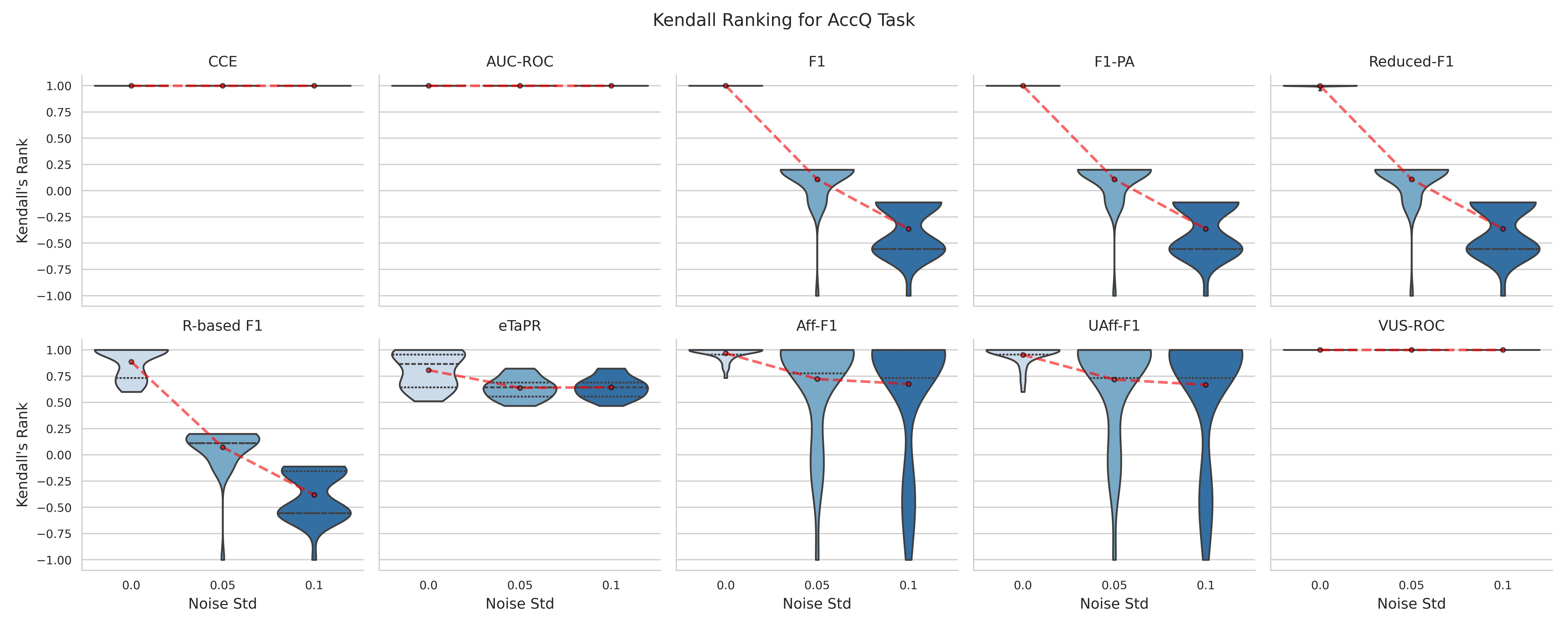}}
   \\
   \subfloat[\label{fig:preq_rank}]{
   \includegraphics[width=1\linewidth]{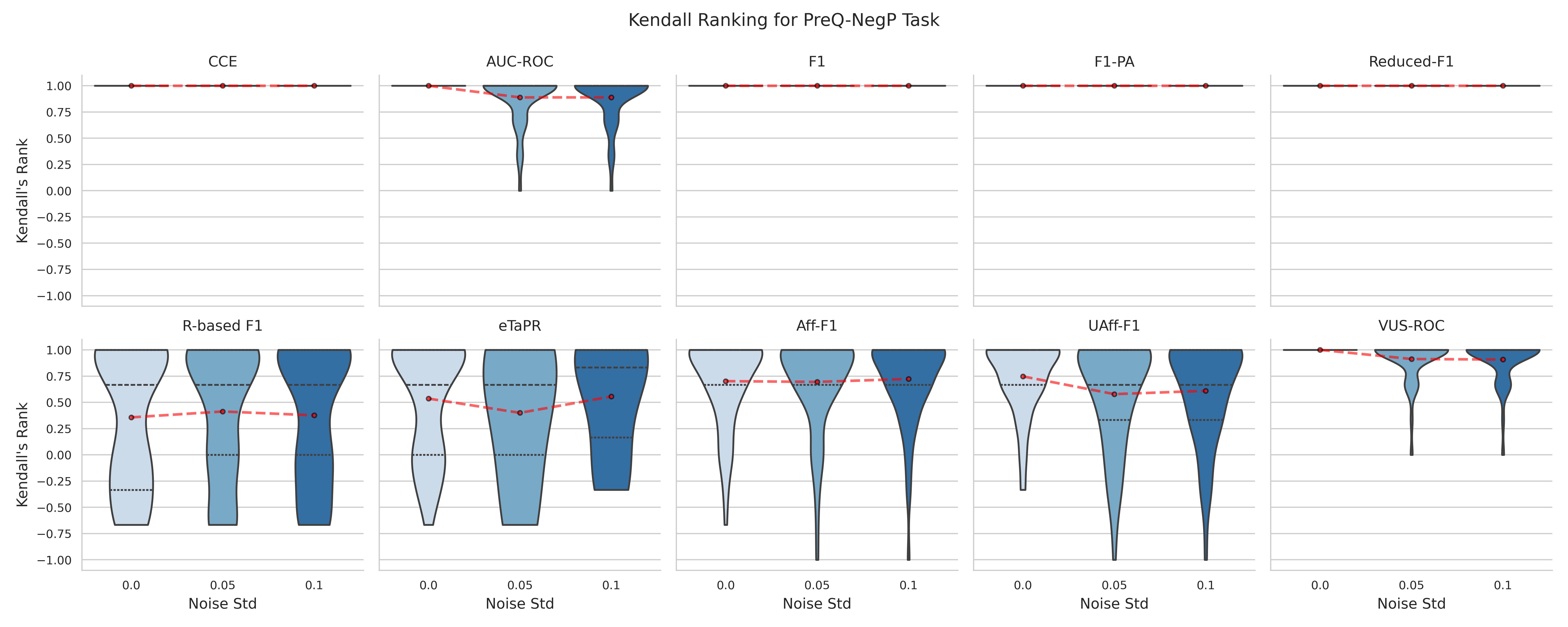}}
   \caption{(a) Performance of different metrics on the AccQ task, (b) Performance of different metrics on the PreQ-NegP task, only considering ranking for \(p\).}
\end{figure}

Figs. \ref{fig:accq_rank} and \ref{fig:preq_rank} respectively demonstrate the performance of different metrics on the AccQ and PreQ-NegP tasks. On the AccQ task, only CCE, AUC-ROC, and VUS-ROC are not affected by noise, while other metrics experience a decline in their ranking capability as noise increases. On the PreQ-NegP task, only CCE, F1, F1-PA, and Reduced-F1 are not affected by noise, while other metrics suffer from incorrect assessment of false positives due to increased noise. Overall, only CCE achieves complete noise robustness, meaning that increasing noise does not affect the metric's ranking capability.

\subsection{Visualization Analysis (\textbf{RQ3})}\label{sec:rq3_vis}

\subsubsection{Synthetic Data Visualization}
\begin{figure}[htb]
   \center
   \includegraphics[width=1\linewidth]{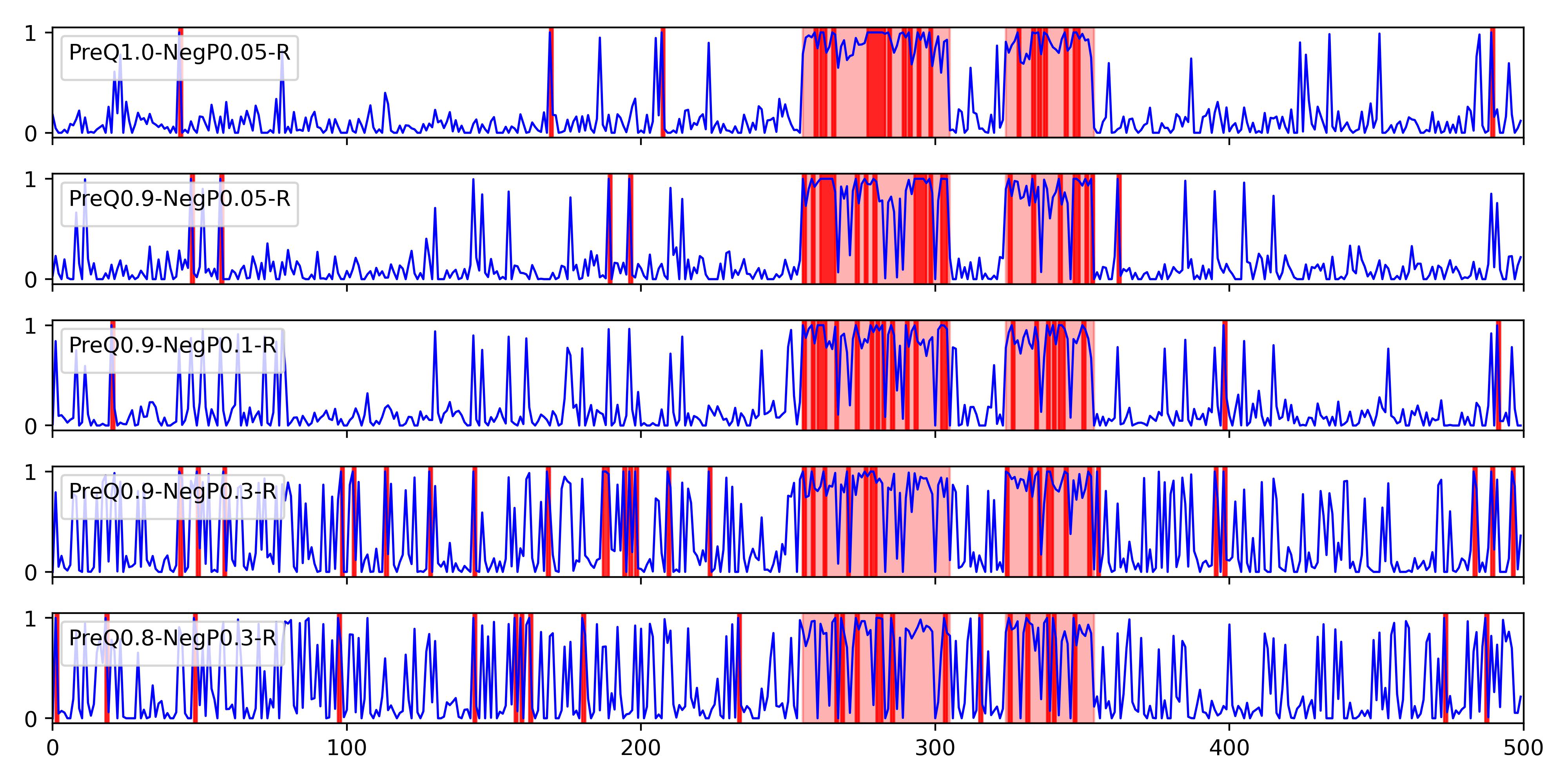}
   \caption{Visualization of different ASGM models on synthetic dataset.}
   \label{fig:syn_plot}

\end{figure}

Figure \ref{fig:syn_plot} shows the visualization results of five PreQ-NegP-R models on synthetic data. Their parameters are shown in the first column of Table \ref{tab:syn_metric}, where the numbers after Q and P represent \(q\) and \(p\) respectively, and the noise size \(\sigma=0.1\). ER represents the theoretical ranking of the model. In this case, only CCE and VUS-ROC can obtain the correct ranking of model performance. Among them, Reduced-F1 has the worst estimation ability, and the score of PreQ0.9-NegP0.1-R, which is expected to rank third, is actually the highest. eTaPR and Aff-F1 also have estimation inaccuracies.


\begin{table}[ht]
   \centering
   \caption{Performance of different ASGM models on synthetic dataset.}
   \resizebox{\linewidth}{!}{
\begin{tabular}{llccccc}
   \toprule
   ER & ASGM/Metric & CCE & Reduced-F1 & eTaPR & Aff-F1 & VUS-ROC \\
   \midrule
   1&PreQ1.0-NegP0.05-R & 0.768 & 0.750 & 0.712 & 0.940 & 0.983 \\
   2&PreQ0.9-NegP0.05-R & 0.707 & 0.706 & 0.706 & 0.957 & 0.974 \\
   3&PreQ0.9-NegP0.1-R & 0.658 & 0.800 & 0.732 & 0.940 & 0.969 \\
   4&PreQ0.9-NegP0.3-R & 0.510 & 0.353 & 0.459 & 0.788 & 0.905 \\
   5&PreQ0.8-NegP0.3-R & 0.412 & 0.480 & 0.516 & 0.811 & 0.888 \\
   \bottomrule
\end{tabular}}
\label{tab:syn_metric}
\end{table}
   
\subsubsection{Real Data Visualization}
\begin{figure}[htb]
   \centering
   \subfloat[ECG example\label{fig:ecg}]{\includegraphics[width=1\linewidth]{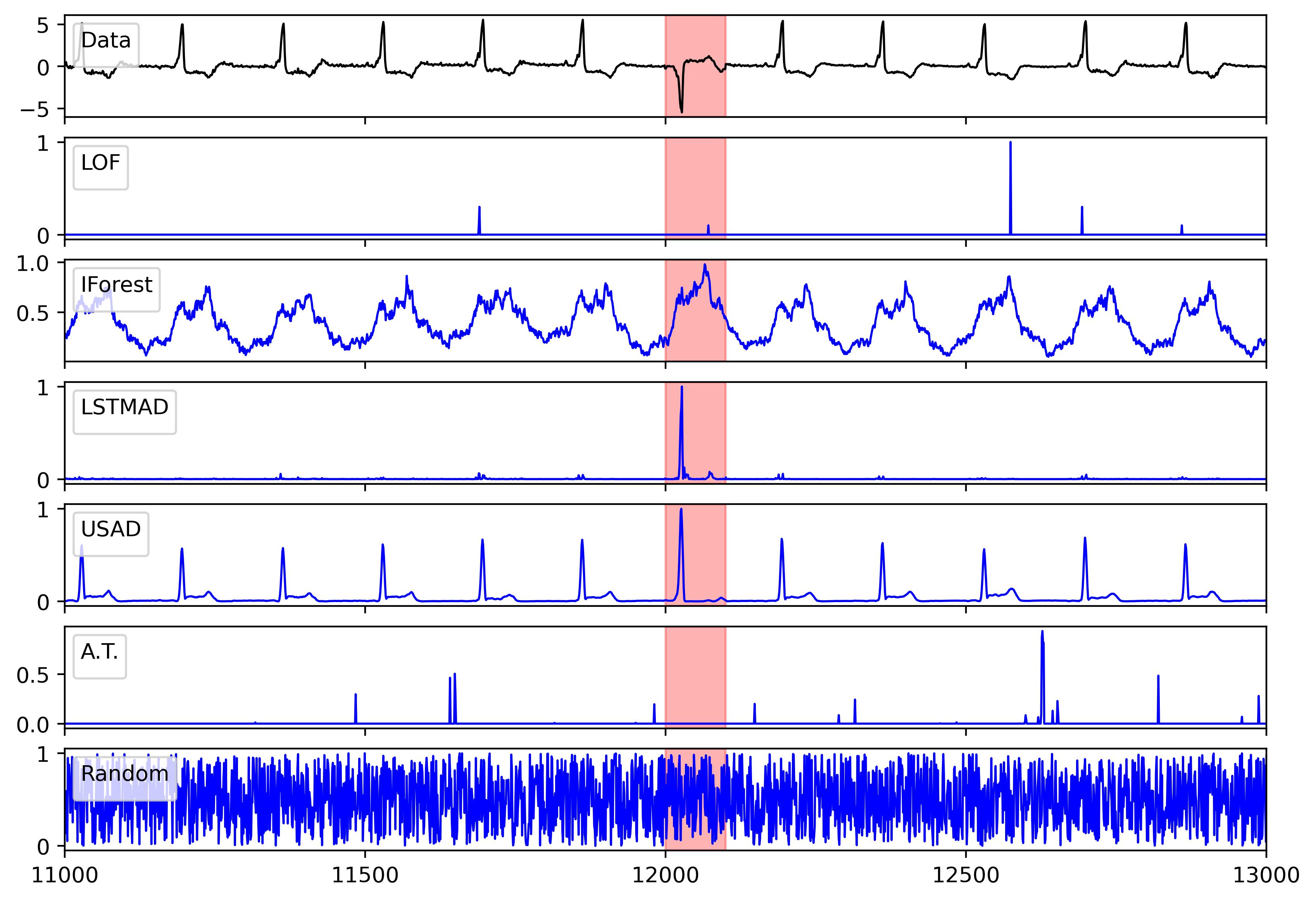}}
   \\
   \subfloat[Power demand example\label{fig:power}]{\includegraphics[width=1\linewidth]{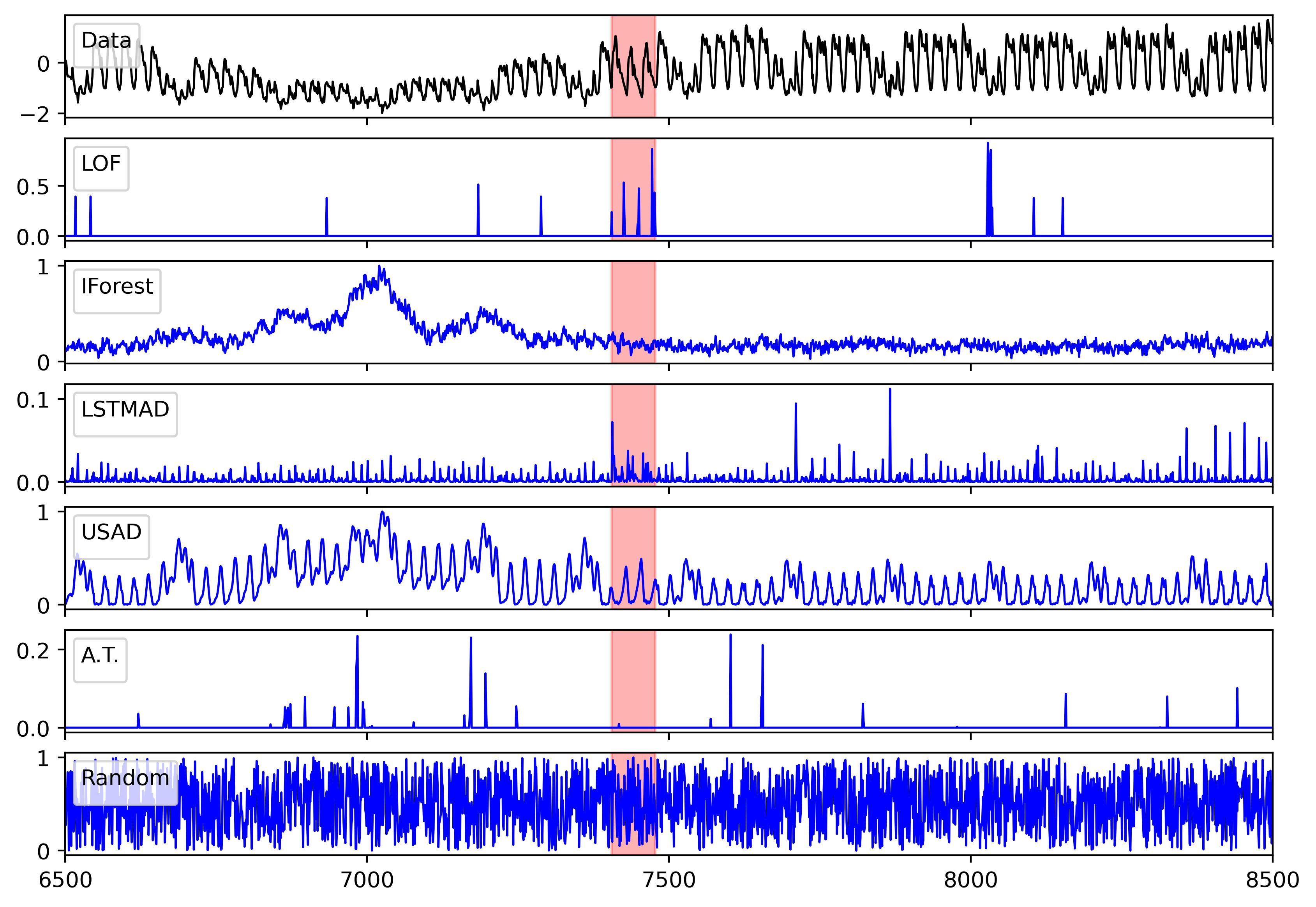}}
   \caption{Visualization of real-world datasets.}
\end{figure}

To better observe and understand anomalies in the dataset, we selected ECG and Power demand from UCR to study the performance of different models in real-world scenarios. The visualization results of model predictions are shown in Figs. \ref{fig:ecg} and \ref{fig:power} respectively. Table \ref{tab:real_metric} shows the scores of these models on different metrics.

In ECG, the models with better performance are LSTMAD, USAD, and A.T. In this case, AUC-ROC performs the worst and cannot accurately evaluate USAD's capability, because AUC-ROC cannot accurately reflect situations where USAD's anomaly detection capability is not prominent enough, as point-based evaluation is too strict. Additionally, Aff-F1's score for LSTMAD is 65.0, which is the lowest among all models, which is not reasonable.

In Power, the models with better performance are LOF, LSTMAD, and IForest, while the anomaly scores of USAD and A.T. in the anomaly region are far below the normal interval. However, at this point, VUS-ROC gives USAD a score of 70.4, considering it a better model, but from the visualization, this model is clearly inferior to LOF, yet their VUS-ROC scores are close.

Furthermore, in both ECG and Power cases, we found that eTaPR always considers the model score to be 0, which weakens the discriminative power of this metric. VUS-ROC gives scores higher than 65 to both Random models, but theoretically, VUS-ROC's scoring for random models should be close to 50. At this point, when the model is close to random, the metric error may exceed 30\%.

\begin{table}[ht]
   \centering
   \caption{Comparison of models on ECG and Power datasets.}
   \resizebox{0.9\linewidth}{!}{
   \begin{tabular}{llccccc}
      \toprule
      ECG & CCE & AUC-ROC & F1 & eTaPR & Aff-F1 & VUS-ROC \\
      \midrule
      LOF & -0.01 & 53.72 & 2.32 & 0.0 & 67.2 & 54.04 \\
      IForest & 22.3 & 80.67 & 4.74 & 3.47 & 69.77 & 82.74 \\
      LSTMAD & 4.66 & 83.53 & 5.33 & 3.57 & 65.0 & 88.19 \\
      USAD & 2.85 & 47.26 & 1.48 & 1.41 & 67.03 & 54.23 \\
      A.T. & -0.21 & 49.27 & 0.0 & 0.0 & 68.46 & 50.26 \\
      Random & 2.82 & 52.85 & 1.04 & 0.0 & 65.42 & 68.49 \\
      \midrule
      Power & CCE & AUC-ROC & F1 & eTaPR & Aff-F1 & VUS-ROC \\
      \midrule
      LOF & 2.63 & 68.56 & 1.83 & 0.0 & 62.65 & 75.45 \\
      IForest & -5.1 & 32.21 & 0.0 & 0.0 & 75.18 & 50.78 \\
      LSTMAD & 0.5 & 64.26 & 2.34 & 2.45 & 67.11 & 79.5 \\
      USAD & -1.75 & 45.31 & 0.78 & 0.0 & 81.8 & 70.4 \\
      A.T. & 0.05 & 51.6 & 1.97 & 0.0 & 71.06 & 54.21 \\
      Random & -4.9 & 45.08 & 0.52 & 0.0 & 67.03 & 65.46 \\
      \bottomrule
   \end{tabular}
   }
   \label{tab:real_metric}
\end{table}

\subsection{Impact of Hyperparameter on CCE (\textbf{RQ4})}\label{sec:rq4_hyper}
The CCE metric has only one important hyperparameter: the confidence threshold \(\tau\). Unlike the buffer size hyperparameters of Aff-F1 and VUS-ROC, this is a dataset-independent hyperparameter. We set the parameter range \(\tau \in [0.1, 0.3, 0.5, 0.7, 0.9]\) and conducted experiments on the AccQ task. Table \ref{tab:cce_hyper} demonstrates the impact of this parameter on CCE's evaluation capability. Clearly, under any \(\tau\) setting, CCE maintains consistent ranking capability, indicating that hyperparameter settings do not affect the absolute ranking of models.

\begin{table}[htbp]
   \centering
   \caption{Sp/Kd/MD scores of CCE under different \(\tau\) settings.}
     \begin{tabular}{c|ccccc}
     \toprule
     \textbf{Threshold} & \textbf{0.1} & \textbf{0.3} & \textbf{0.5} & \textbf{0.7} & \textbf{0.9} \\
     \midrule
     \textbf{\(\tau\)} & 1/1/0 & 1/1/0 & 1/1/0 & 1/1/0 & 1/1/0 \\
     \bottomrule
     \end{tabular}%
   \label{tab:cce_hyper}%
 \end{table}%

\section{Conclusion}
This paper addresses the limitations of previous TSAD evaluation metrics by proposing the CCE metric that characterizes confidence and uncertainty consistency, solving issues such as hyperparameter dependency, low robustness, high computational cost, and lack of prediction consistency evaluation in previous metrics. Additionally, the RankEval benchmark for metric evaluation was constructed, which can assess metric performance across different tasks and robust scenarios. Through theoretical analysis, the boundedness, linear complexity, and robustness of the proposed CCE were proven. Furthermore, extensive numerical experiments validated these properties and hyperparameter independence, while demonstrating the robustness limitations of other metrics. Additionally, combined with visualization analysis, the effectiveness and robustness of CCE were further demonstrated. 
The CCE metric and RankEval benchmark provide a comprehensive TSAD model evaluation solution, advancing TSAD methodologies and facilitating reliable practical model selection.

\section*{Acknowledgment}
This work was supported in part by National Natural Science Foundation of China No. 92467109, U21A20478, National Key R\&D Program of China 2023YFA1011601, and the Major Key Project of PCL, China under Grant PCL2025AS11.

\bibliography{main}
\bibliographystyle{IEEEtran}

\begin{IEEEbiography}[{\includegraphics[width=1in,height=1.25in,clip,keepaspectratio]{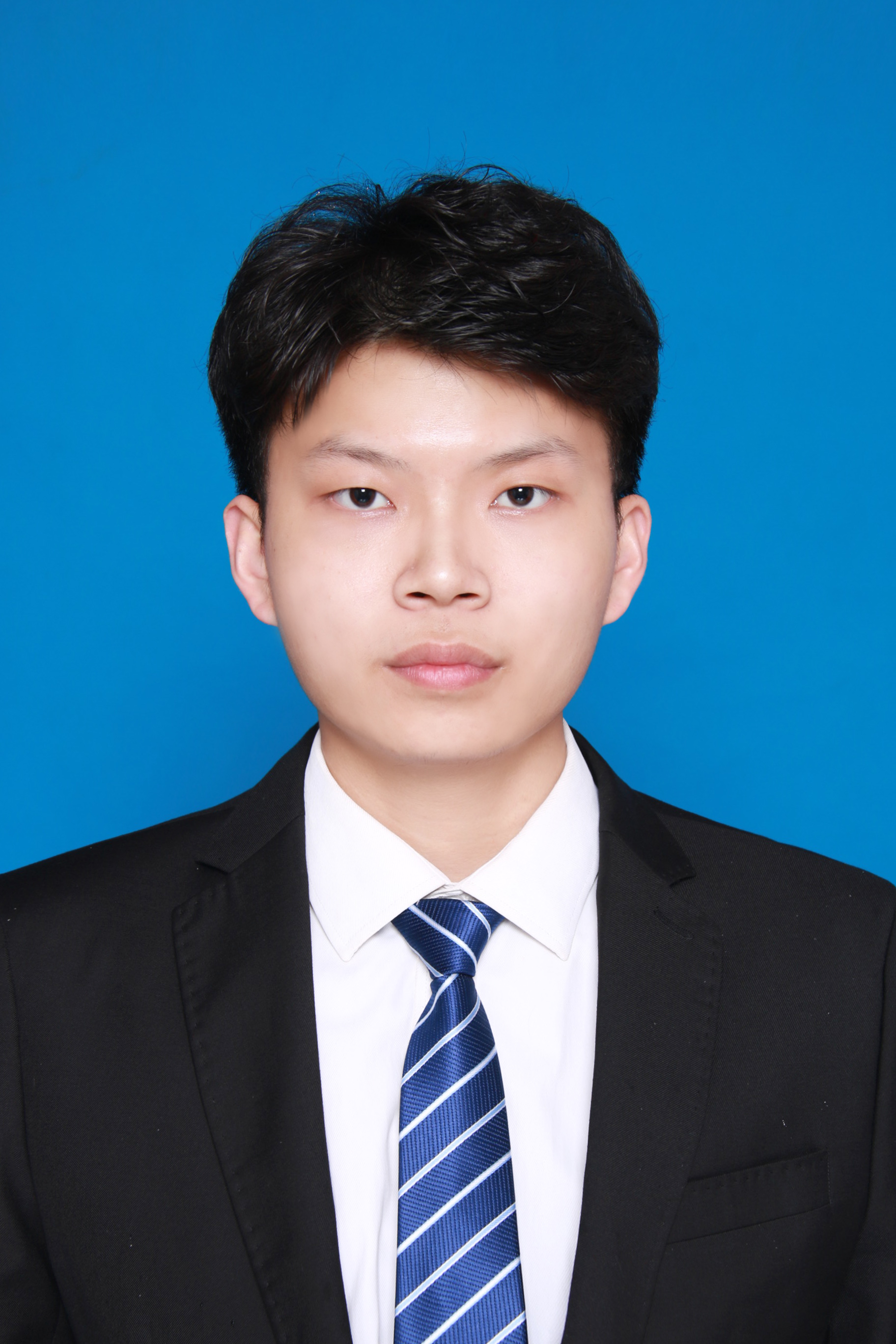}}]{Zhijie Zhong} received the B.S. degree in 2022 from the Harbin Engineering University, Harbin, China and he is currently pursuing the Ph.D. degree in the School of Future Technology, South China University of Technology, Guangzhou, China. His research interests include data mining, machine learning, time series analysis, anomaly detection, and large language model (LLM).
\end{IEEEbiography}
\begin{IEEEbiography} [{\includegraphics[width=1in,height=1.25in,clip,keepaspectratio]{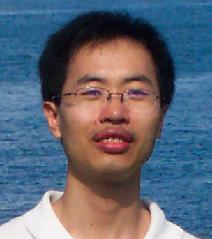}}]{Zhiwen Yu (S'06-M'08-SM'14)} is a Professor in School of Computer Science and Engineering, South China University of Technology, China. He received the Ph.D. degree from the City University of Hong Kong, Hong Kong, in 2008. Dr. Yu has authored or coauthored more than 200 refereed journal articles and international conference papers, including more than 70 articles in the journals of IEEE Transactions. His google citation is more than 10000, and h-index is 44. He is an Associate Editor of the IEEE Transactions on systems, man, and cybernetics: systems. He is a senior member of IEEE and ACM, a Member of the Council of China Computer Federation (CCF).
\end{IEEEbiography}
\begin{IEEEbiography}[{\includegraphics[width=1in,height=1.25in,clip,keepaspectratio]{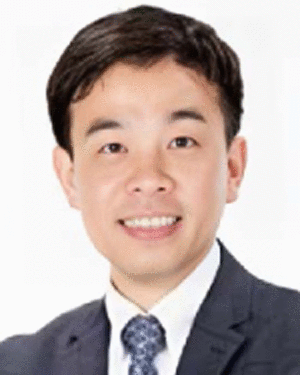}}] {Yiu-Ming Cheung (Fellow, IEEE)} received the PhD degree from the Department of Computer Science and Engineering, The Chinese University of Hong Kong, Hong Kong, in 2000. He is currently a chair professor with the Department of Computer Science, Hong Kong Baptist University, Hong Kong. His current research interests include machine learning, pattern recognition, and visual computing. He is a fellow of the American Association for the Advancement of Science (AAAS), Institution of Engineering and Technology (IET), British Computer Society (BCS), and Asia-Pacific Artificial Intelligence Association (AAIA). He is the editor-in-chief of IEEE Transactions on Emerging Topics in Computational Intelligence. He also served as an associate editor for IEEE Transactions on Cybernetics, IEEE Transactions on Cognitive and Developmental Systems, IEEE Transactions on Neural Networks and Learning Systems from 2014 to 2020, Pattern Recognition, Knowledge and Information Systems, and Neurocomputing, just to name a few.
\end{IEEEbiography}
\begin{IEEEbiography} [{\includegraphics[width=1in,height=1.25in,clip,keepaspectratio]{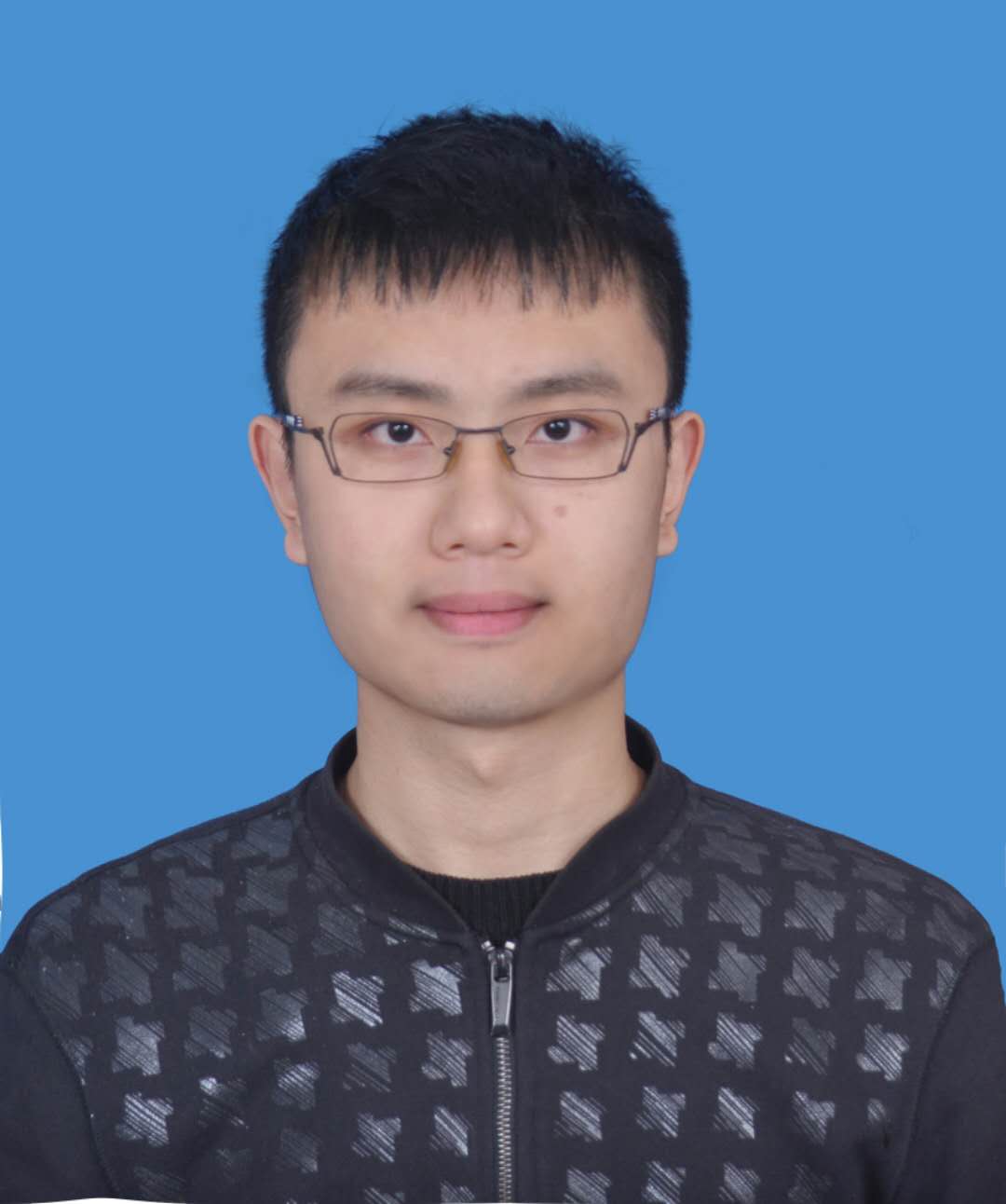}}] {Kaixiang Yang (M'21)} received the B.S. degree and M.S. degree from the University of Electronic Science and Technology of China and Harbin Institute of Technology, China, in 2012 and 2015, respectively, and the Ph.D. degree from the School of Computer Science and Engineering, South China University of Technology, China, in 2020. He has been a Research Engineer with the 7th Research Institute, China Electronics Technology Group Corporation, Guangzhou, China, from 2015 to 2017, and has been a Postdoctoral Researcher with Zhejiang University from 2020 to 2021. He is now with the School of Computer Science and Engineering, South China University of Technology. His research interests include pattern recognition, machine learning, and industrial data intelligence.
\end{IEEEbiography}

\ifarxiv
  \clearpage
  \newcommand{\includedfrommain}{} 
\ifx\includedfrommain\undefined
    \documentclass[lettersize,journal]{IEEEtran}
    \usepackage{xr-hyper}
    \usepackage{hyperref}
    \usepackage{amsmath,amsfonts}
    \usepackage{algorithmic}
    \usepackage{algorithm}
    \usepackage{array}
    \usepackage[caption=false,font=normalsize,labelfont=sf,textfont=sf]{subfig}
    \usepackage{textcomp}
    \usepackage{stfloats}
    \usepackage{url}
    \usepackage{verbatim}
    \usepackage{graphicx}

    \usepackage{pifont}
    
    \usepackage{hyperref}
    \usepackage{url}
    
    \usepackage{booktabs}       
    \usepackage{nicefrac}       
    \usepackage{microtype}      
    \usepackage{xcolor}         
    \usepackage{graphicx}
    
    \usepackage[normalem]{ulem} 
    \usepackage{multirow}
    \usepackage{multicol}
    \usepackage{wrapfig}
    
    \usepackage[scr=boondox,cal=esstix]{mathalpha}
    \usepackage{float}                  
    \usepackage{overpic}                
    \usepackage{listings}
    \usepackage[numbers]{natbib}
    
    \usepackage{xr}
    \externaldocument{output}
    
    \usepackage{xcite}
    \externalcitedocument{output}
    
    \newcommand{\rv}[1]{#1}
    \lstset{
    language=Python,  
    basicstyle=\ttfamily,  
    tabsize=4,  
    numbers=left,  
    numberstyle=\tiny\color{gray},  
    breaklines=true,  
    frame=single,  
    stringstyle  = \color{purple},
    keywordstyle = \color{blue!60!black}\bfseries,
    commentstyle = \color{green!40!black}\scshape,
    xleftmargin      = 20pt,
    xrightmargin     = 0pt,
    aboveskip=5pt,
    belowskip=-2.2pt,
    frame            = tb,
    framesep         = \fboxsep,
    framexleftmargin = 20pt,
    breaklines=true,
}

  \externaldocument{main} 
  \begin{document}
\fi


\renewcommand{\thetable}{S\arabic{table}}
\renewcommand{\thefigure}{S\arabic{figure}}
\renewcommand{\thealgorithm}{S\arabic{algorithm}}

\hyphenation{op-tical net-works semi-conduc-tor IEEE-Xplore}

\setcounter{section}{0}
\renewcommand{\thesection}{\Alph{section}}


\section*{Appendix}

This is the appendix of \textit{CCE: Confidence-Consistency Evaluation for Time Series Anomaly Detection}.

\section{Supplementary Experiments}

\subsection{Datasets}
Table \ref{tab:metric_ana_config} illustrates the characteristics of all datasets employed in RankEval, which encompasses real-world datasets: MSL, SMD, SWAT, Creditcard, and PSM \cite{PSM,MSL,SWAT}.

A total of 30 synthetic datasets were generated, with examples named \textit{100k-20seg-50L} and \textit{100k-20seg-50H}. In such names, \textit{100k} refers to the total length of the time series (corresponding to the TS Length column), \textit{20seg} means that 20 anomaly segments are expected to be generated (i.e., the \textit{Segments} column), and \textit{50L} indicates that the average length of anomaly segments is 50 with low variance (for instance, lengths ranging from 40 to 60). In contrast, \textit{50H} also represents an average segment length of 50 but with higher variance (such as lengths between 1 and 99). The \textit{Max/Min Seg Length} columns respectively stand for the maximum and minimum lengths of the generated anomaly segments. For time series lengths of 100k and 10k, 15 synthetic datasets were created each. Moreover, 6 real-world datasets were selected according to their characteristics like length, number of anomalies, and anomaly lengths.

\begin{table}[htbp]
    \centering
    \caption{Time series characteristics of synthetic and real data.}
    \resizebox{\linewidth}{!}{
      \begin{tabular}{l|rrrr}
      \toprule
      Dataset Name & {TS Length} & {Segments} & {Max Seq Length} & {Min Seq Length} \\
      \midrule
      100k-20seg-50L & 100000 & 20    & 60    & 40 \\
      100k-200seg-50L & 100000 & 200   & 60    & 40 \\
      100k-20seg-50H & 100000 & 20    & 99    & 1 \\
      100k-200seg-50H & 100000 & 200   & 99    & 1 \\
      100k-50seg-20L & 100000 & 50    & 30    & 10 \\
      100k-500seg-20L & 100000 & 500   & 30    & 10 \\
      100k-50seg-20H & 100000 & 50    & 39    & 1 \\
      100k-500seg-20H & 100000 & 500   & 39    & 1 \\
      100k-10seg-100L & 100000 & 10    & 110   & 90 \\
      100k-100seg-100L & 100000 & 100   & 110   & 110 \\
      100k-10seg-100H & 100000 & 10    & 199   & 1 \\
      100k-100seg-100H & 100000 & 100   & 199   & 1 \\
      100k-2seg-500L & 100000 & 2     & 550   & 450 \\
      100k-20seg-500L & 100000 & 20    & 550   & 450 \\
      100k-2seg-500H & 100000 & 2     & 999   & 1 \\
      100k-20seg-500H & 100000 & 20    & 999   & 1 \\
      \midrule
      10k-2seg-50L & 10000 & 2     & 60    & 40 \\
      10k-20seg-50L & 10000 & 20    & 60    & 40 \\
      10k-2seg-50H & 10000 & 2     & 99    & 1 \\
      10k-20seg-50H & 10000 & 20    & 99    & 1 \\
      10k-5seg-20L & 10000 & 5     & 30    & 10 \\
      10k-50seg-20L & 10000 & 50    & 30    & 10 \\
      10k-5seg-20H & 10000 & 5     & 39    & 1 \\
      10k-50seg-20H & 10000 & 50    & 39    & 1 \\
      10k-1seg-100L & 10000 & 1     & 110   & 90 \\
      10k-10seg-100L & 10000 & 10    & 110   & 110 \\
      10k-1seg-100H & 10000 & 1     & 199   & 1 \\
      10k-10seg-100H & 10000 & 10    & 199   & 1 \\
      10k-2seg-500L & 10000 & 2     & 550   & 450 \\
      10k-2seg-500H & 10000 & 2     & 999   & 1 \\
      \midrule
      MSL  & 73729 & 36    & 1141  & 11 \\
      Creditcard & 284807 & 465   & 5     & 1 \\
      SWAT  & 449919 & 35    & 35900 & 101 \\
      SMD-1-1 & 28479 & 8     & 721   & 2 \\
      SMD-2-1 & 23694 & 13    & 452   & 8 \\
      SMD-3-1 & 28700 & 4     & 131   & 21 \\
      PSM & 87841 & 72 & 8861 & 1 \\
      \bottomrule
      \end{tabular}}
    \label{tab:metric_ana_config}%
  \end{table}

\subsection{Model Comparison in Real-World Scenario}
Table \ref{tab:real_world_model_comparison} illustrates the average performance of multiple TSAD models across the MSL, SWAT, PSM, SMD, and Creditcard datasets. Current analysis reveals that no single state-of-the-art model achieves superior performance across all evaluation metrics. Additionally, the consistently low CCE scores across these models suggest that current approaches have not sufficiently considered the prediction consistency issue. This observation highlights a promising new direction for advancing time series anomaly detection models in future research.

\begin{table}[ht]
    \centering
    \caption{Comparative performance of different time series models in real-world scenario (results are averaged).}
    \resizebox{\linewidth}{!}{
        \begin{tabular}{l|llllll}
            \toprule
            Avg. & CCE & AUC-ROC & F1 & eTaPR & Aff-F1 & VUS-ROC \\
            \midrule
            LOF & 0.27 & 55.25 & 8.04 & 7.15 & 69.69 & 68.32 \\
            IForest & 2.1 & 54.58 & 9.98 & 8.83 & 49.81 & 67.22 \\
            LSTMAD & 1.46 & 79.87 & 25.5 & 18.53 & 61.63 & 84.71 \\
            USAD & 1.77 & 76.84 & 23.64 & 16.31 & 69.51 & 81.54 \\
            A.T. & 0.25 & 50.3 & 4.66 & 3.49 & 62.81 & 58.79 \\
            Donut & 0.23 & 71.07 & 19.08 & 14.37 & 61.26 & 79.55 \\
            TimesNet & 0.22 & 58.52 & 12.54 & 12.62 & 73.68 & 71.55 \\
            \bottomrule
            \end{tabular}}
    \label{tab:real_world_model_comparison}
    \end{table}

\subsection{Robustness Analysis}
\begin{figure*}[ht]
    \centering
    \subfloat[]{\includegraphics[width=0.33\linewidth]{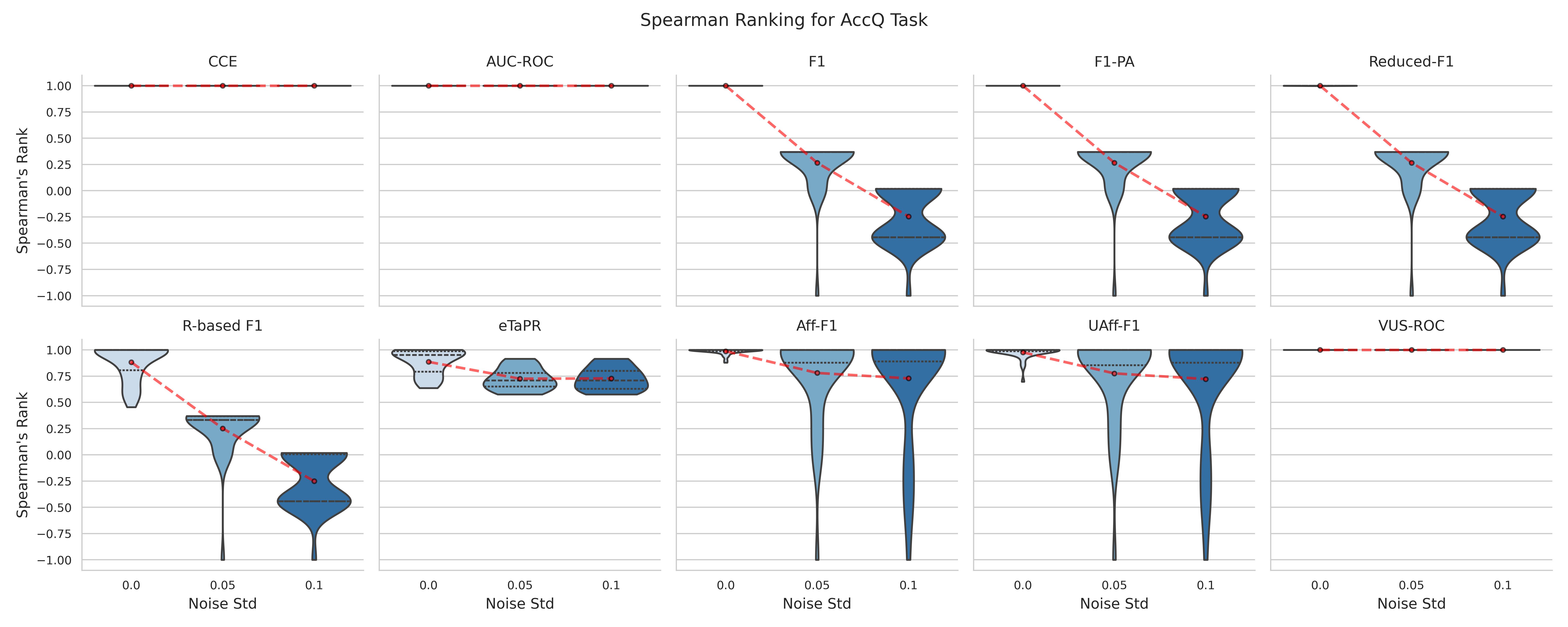}}
    \subfloat[]{\includegraphics[width=0.33\linewidth]{imgs/AccQ_rank_Kd.jpg}}
    \subfloat[]{\includegraphics[width=0.33\linewidth]{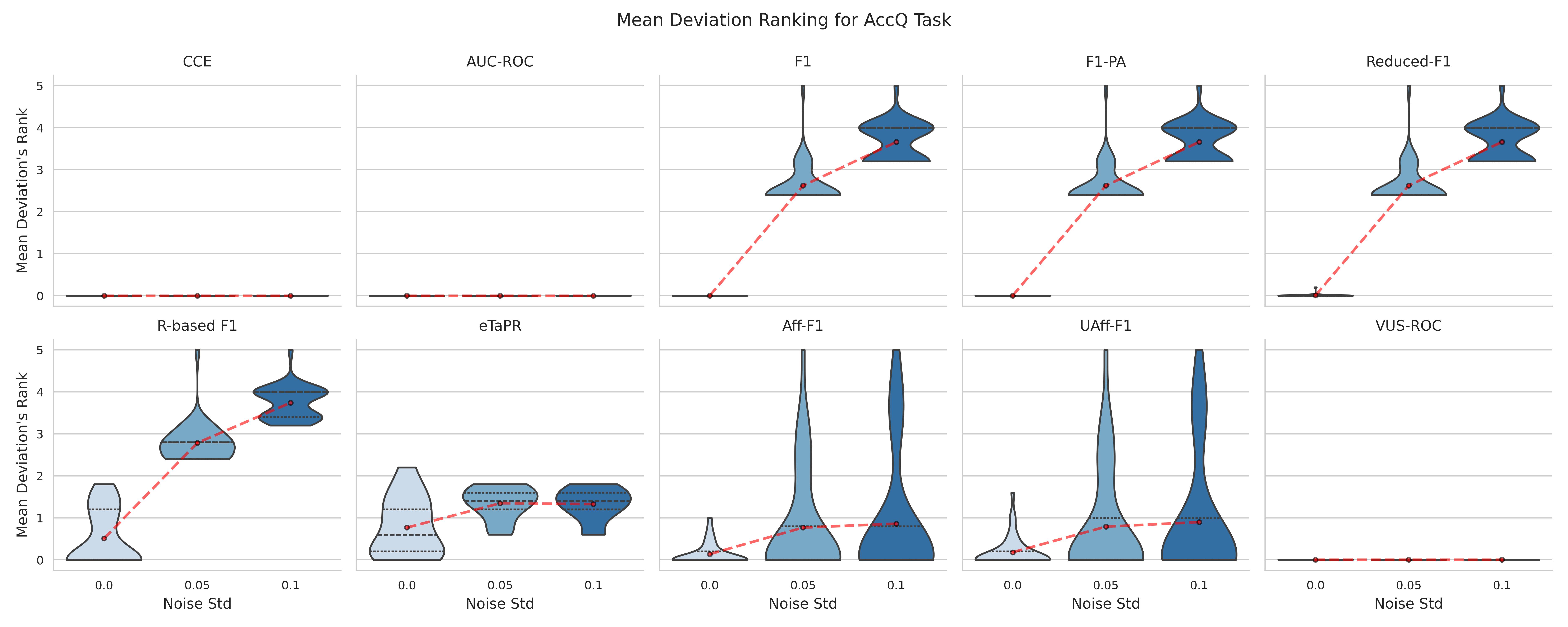}}
    \caption{Evaluation of ranking capability of different metrics on AccQ task. (a) Spearman correlation (b) Kendall correlation (c) Mean Rank Deviation}
    \label{fig:accq_all}
\end{figure*}

\begin{figure*}[ht]
  \centering
  \subfloat[]{\includegraphics[width=0.33\linewidth]{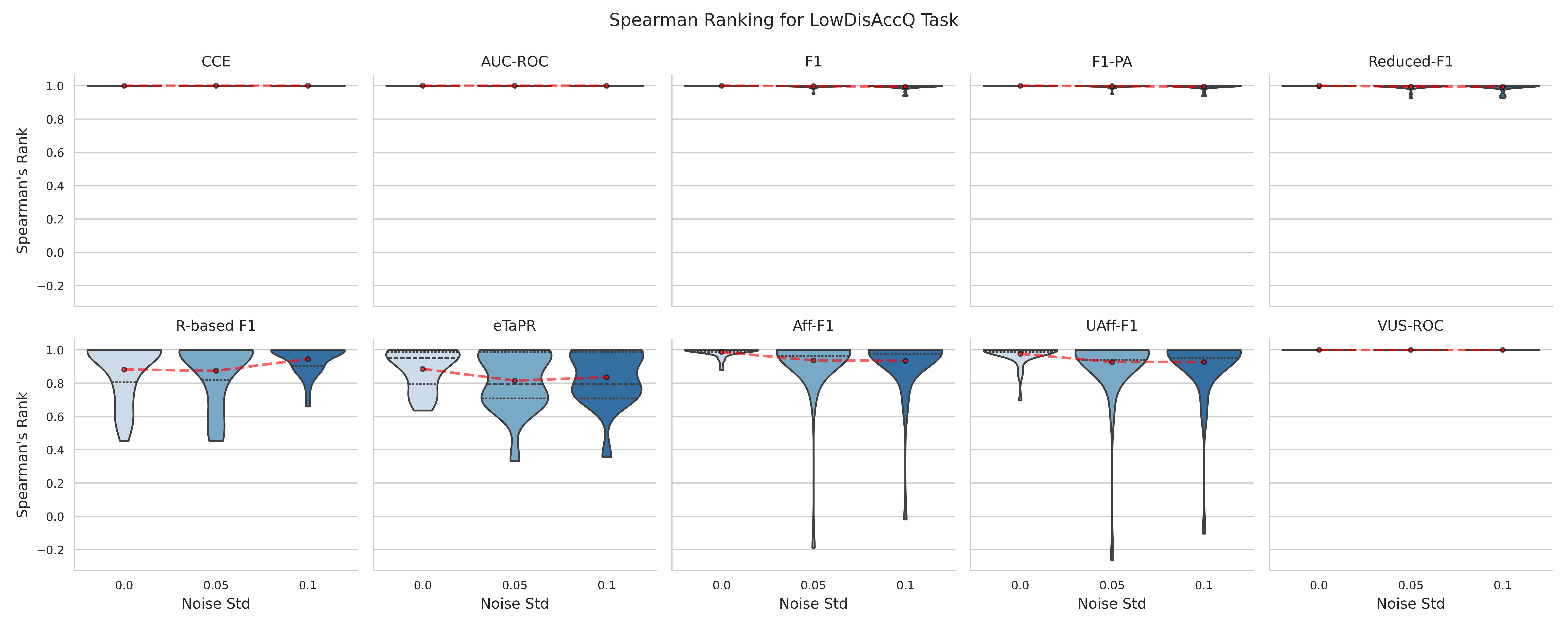}}
  \subfloat[]{\includegraphics[width=0.33\linewidth]{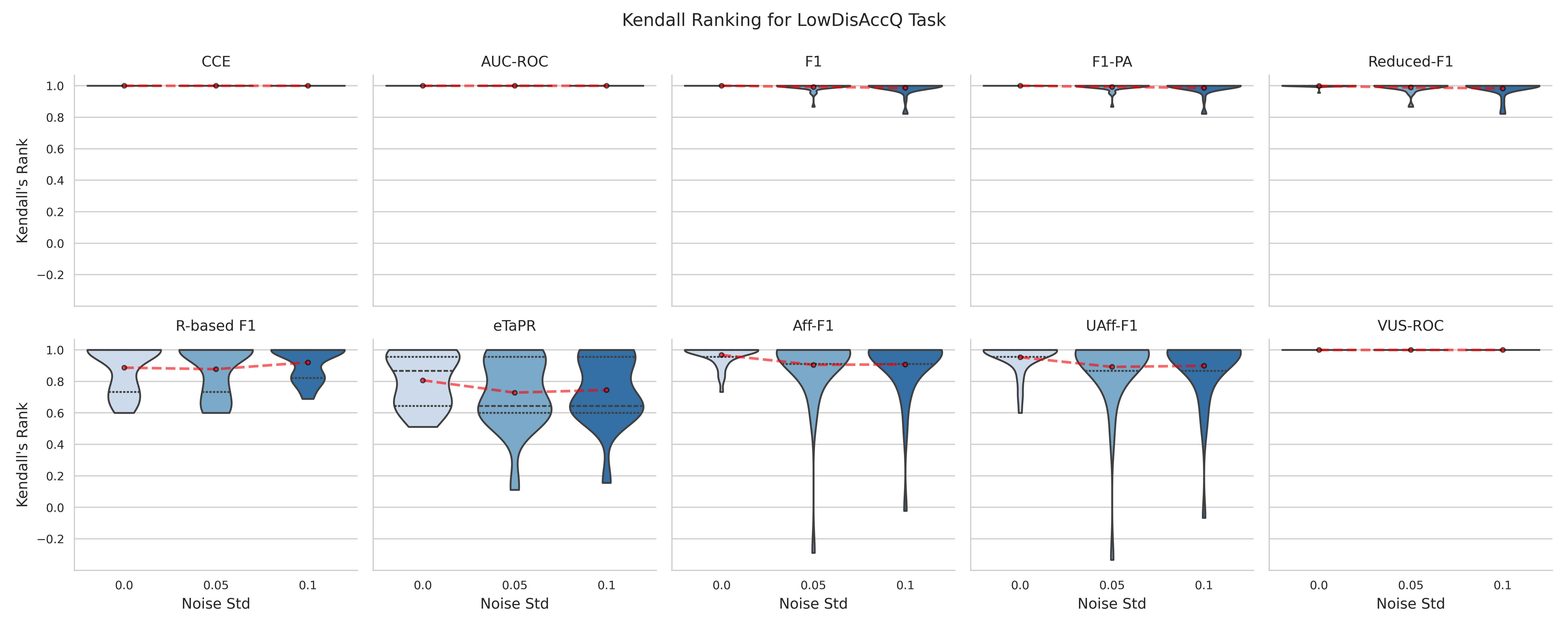}}
  \subfloat[]{\includegraphics[width=0.33\linewidth]{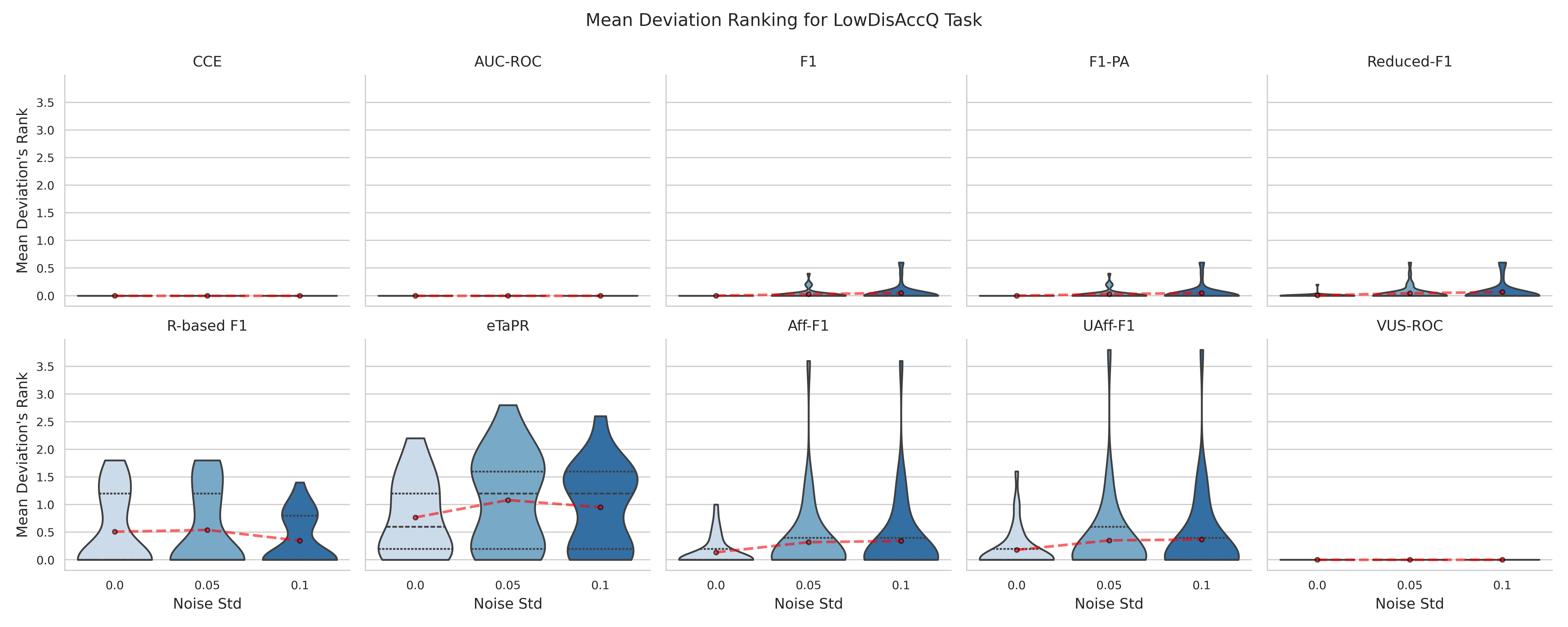}}
  \caption{Evaluation of ranking capability of different metrics on LowDisAccQ task. (a) Spearman correlation (b) Kendall correlation (c) Mean Rank Deviation}
  \label{fig:lowdisaccq_all}
\end{figure*}

\begin{figure*}[ht]
  \centering
  \subfloat[]{\includegraphics[width=0.33\linewidth]{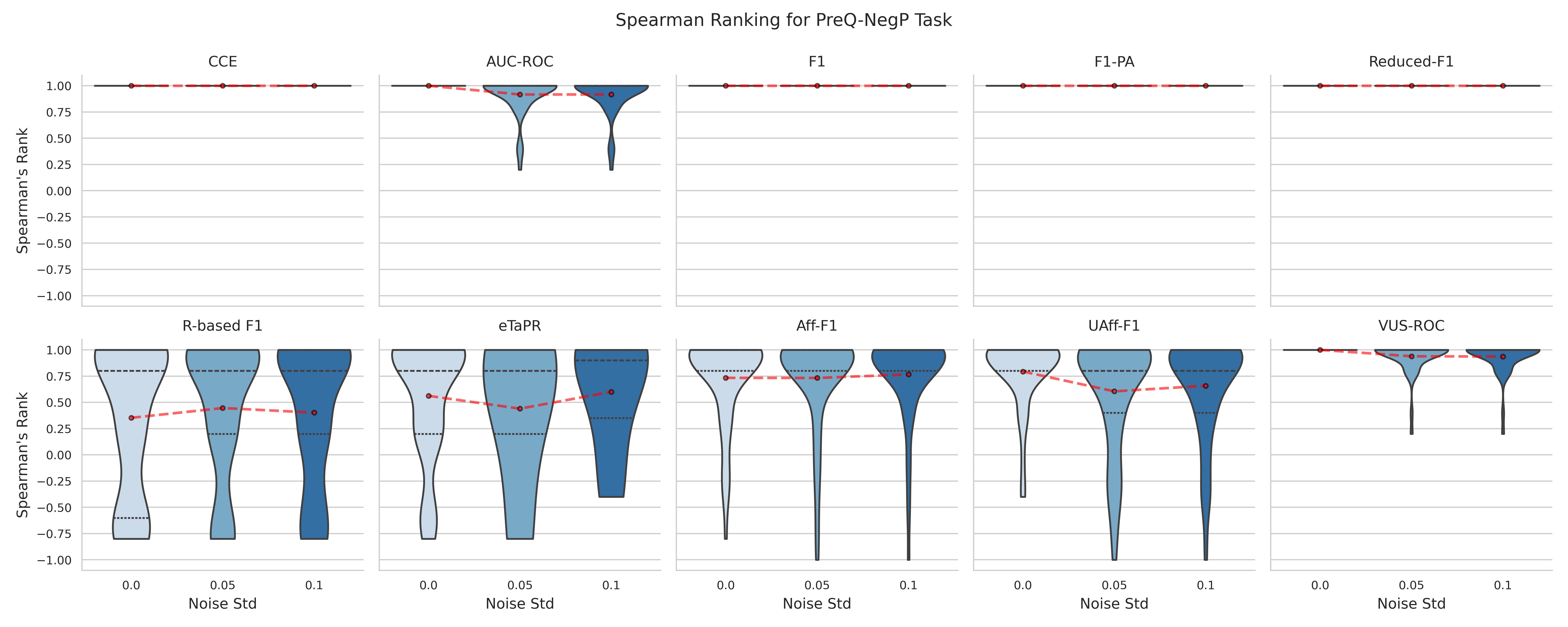}}
  \subfloat[]{\includegraphics[width=0.33\linewidth]{imgs/PreQ-NegP_p_rank_Kd.jpg}}
  \subfloat[]{\includegraphics[width=0.33\linewidth]{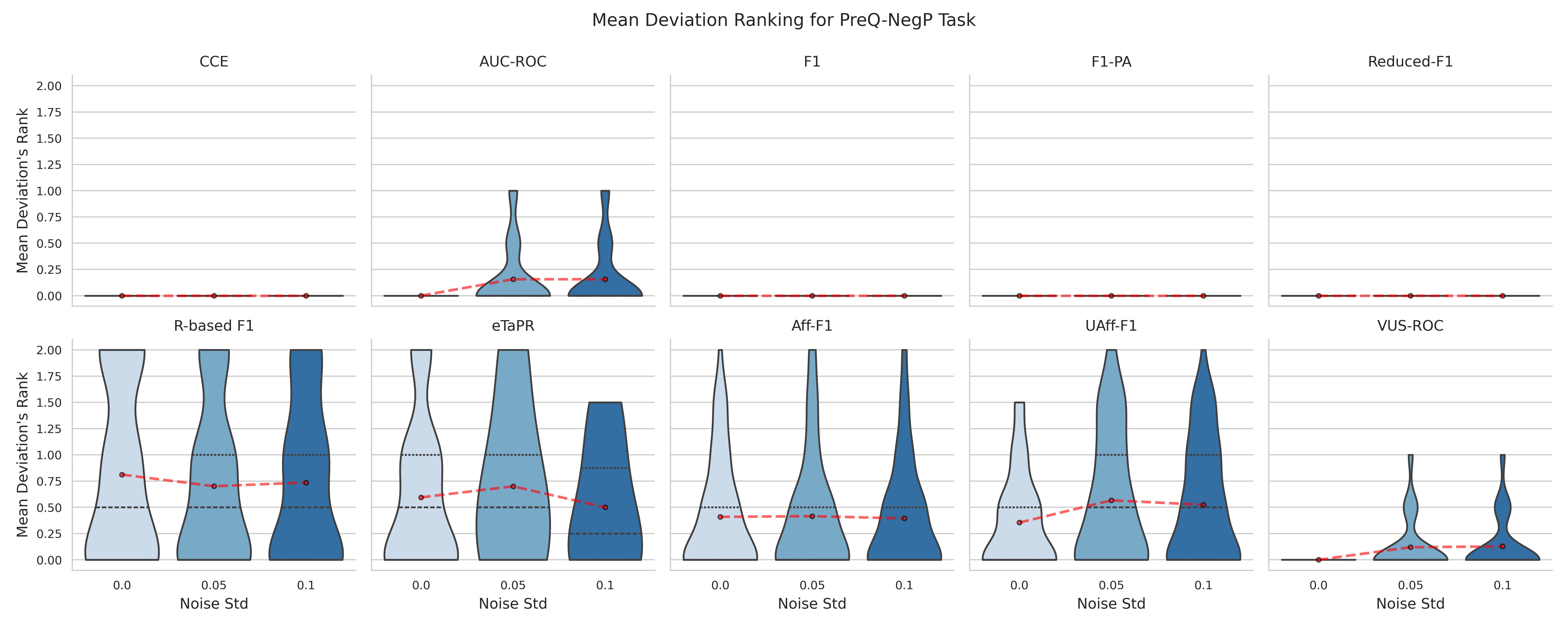}}
  \caption{Evaluation of ranking capability of different metrics on PreQ-NegP task, only considering ranking for \(p\). (a) Spearman correlation (b) Kendall correlation (c) Mean Rank Deviation}
  \label{fig:preq_negpp_all}
\end{figure*}

\begin{figure*}[ht]
  \centering
  \subfloat[]{\includegraphics[width=0.33\linewidth]{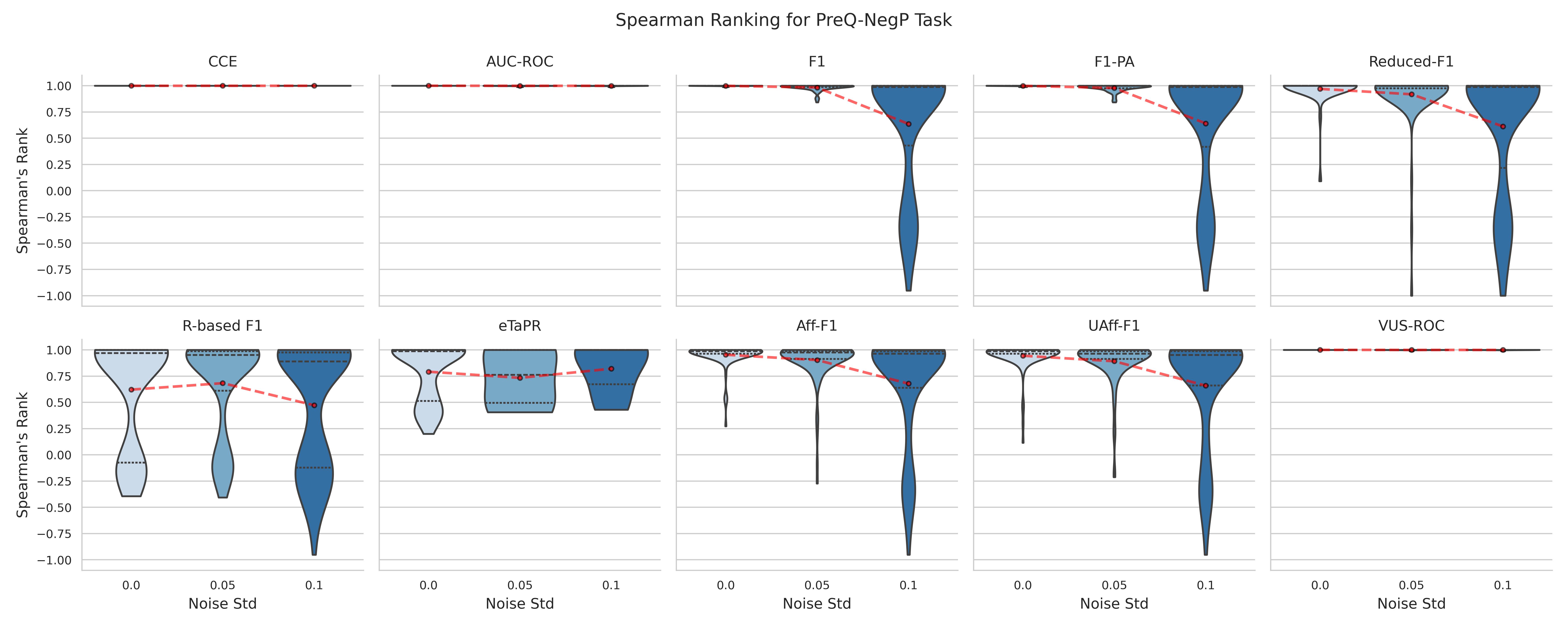}}
  \subfloat[]{\includegraphics[width=0.33\linewidth]{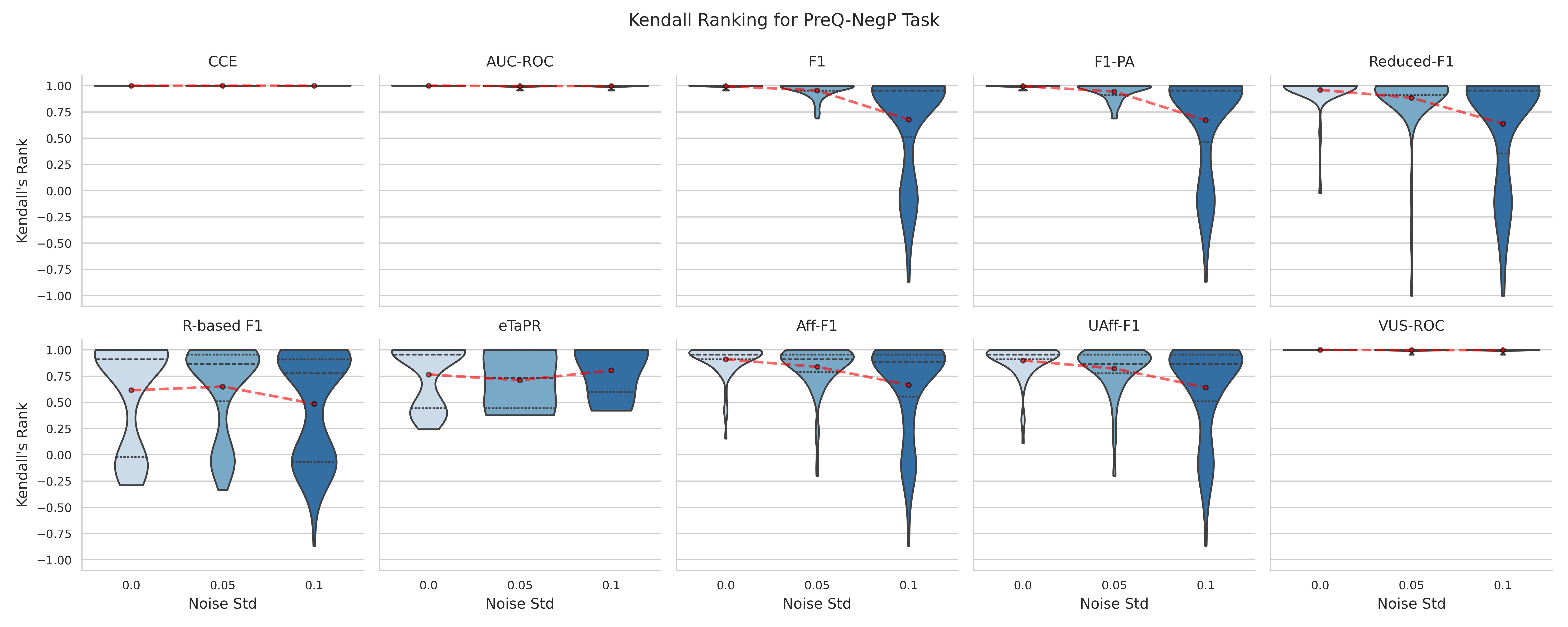}}
  \subfloat[]{\includegraphics[width=0.33\linewidth]{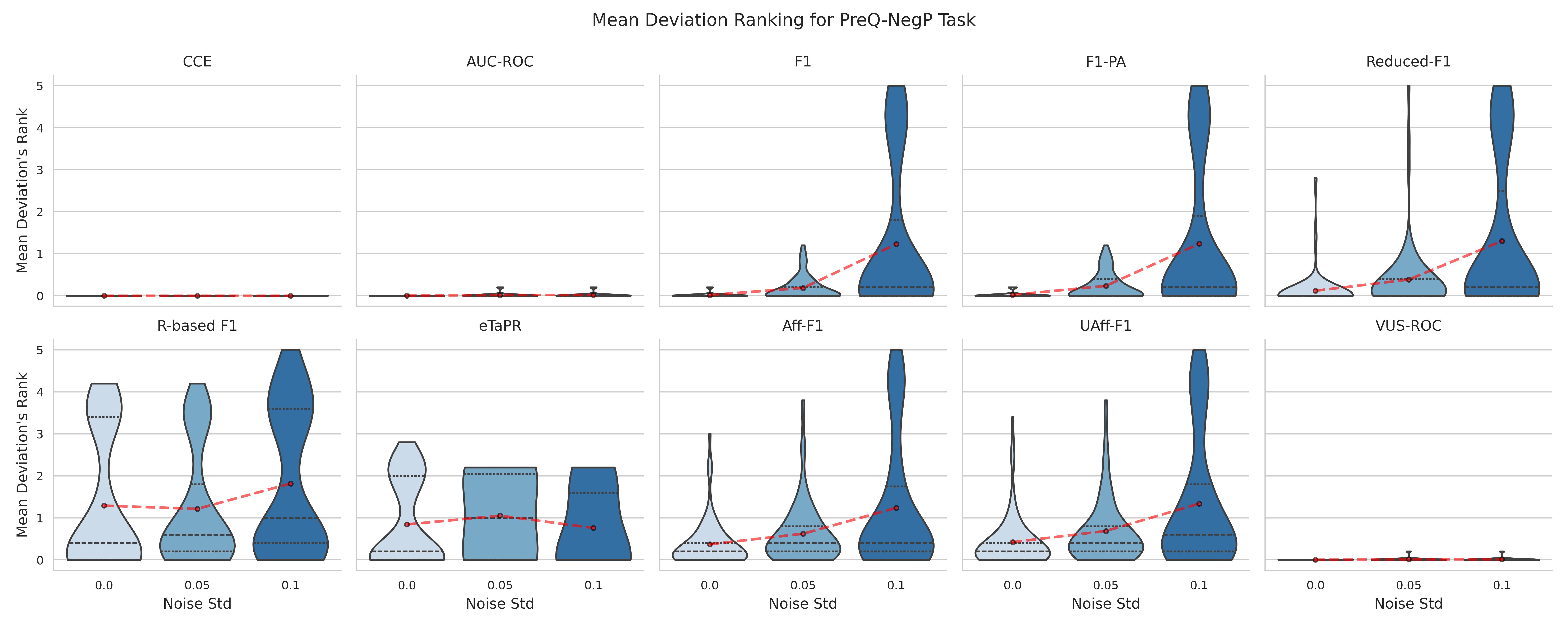}}
  \caption{Evaluation of ranking capability of different metrics on PreQ-NegP task, only considering ranking for \(q\). (a) Spearman correlation (b) Kendall correlation (c) Mean Rank Deviation}
  \label{fig:preq_negpq_all}
\end{figure*}

Figs. \ref{fig:accq_all} and \ref{fig:lowdisaccq_all} respectively demonstrate the sensitivity of different metrics to noise on the AccQ and LowDisAccQ tasks. Figs. \ref{fig:preq_negpp_all} and \ref{fig:preq_negpq_all} both show the sensitivity of different metrics to noise on the PreQ-NegP task, separately considering only the effects of \(p\) and \(q\) respectively. We found that only CCE can maintain robustness to noise across all tasks, while other metrics such as F1, F1-PA, and Reduced-F1 cannot maintain robustness on AccQ and PreQ-NegP tasks. R-based F1, eTaPR, Aff-F1, and UAff-F1 cannot maintain robustness on AccQ, LowDisAccQ, and PreQ-NegP tasks, and even under noise-free conditions, they still cannot achieve perfect ranking, indicating that these metrics may be prone to inflated evaluation. Additionally, AUC-ROC and VUS-ROC show slightly insufficient robustness when considering \(q\), but their performance on other tasks is far better than other comparison metrics.

\subsection{Impact of Noise on Different Metrics}\label{app:noise_effect}
\begin{figure*}
\subfloat[CCE\label{fig:robust_cce}]{\includegraphics[width=0.5\linewidth]{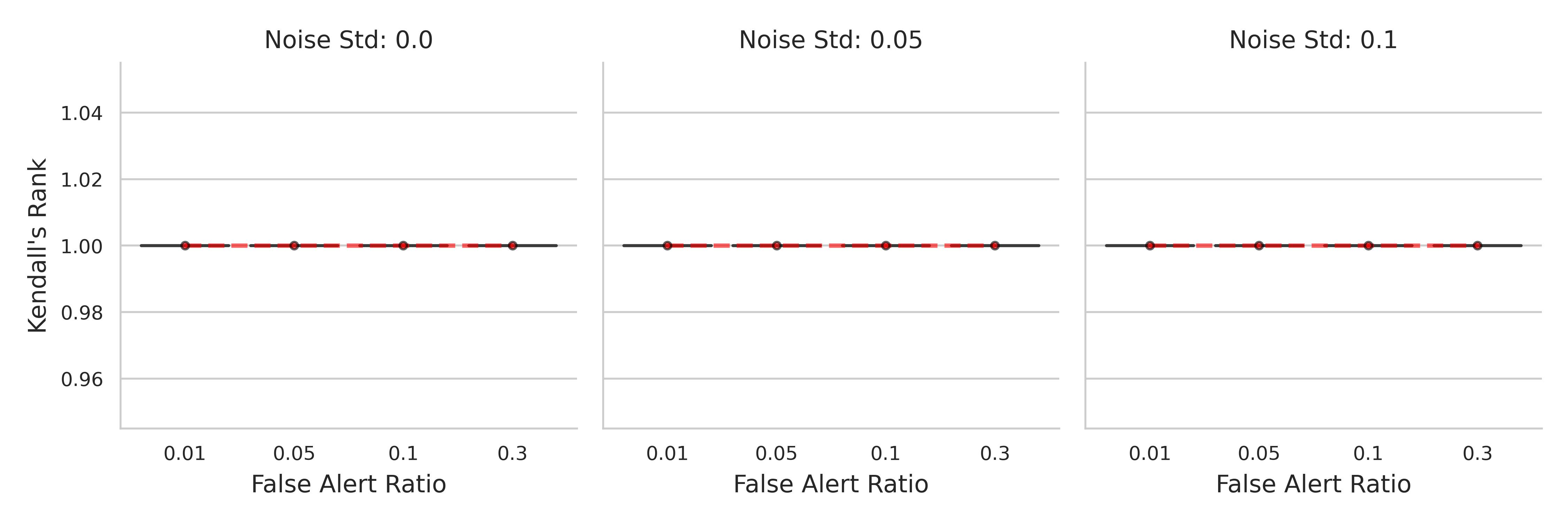}}
\subfloat[Aff-F1\label{fig:robust_aff-f1}]{\includegraphics[width=0.5\linewidth]{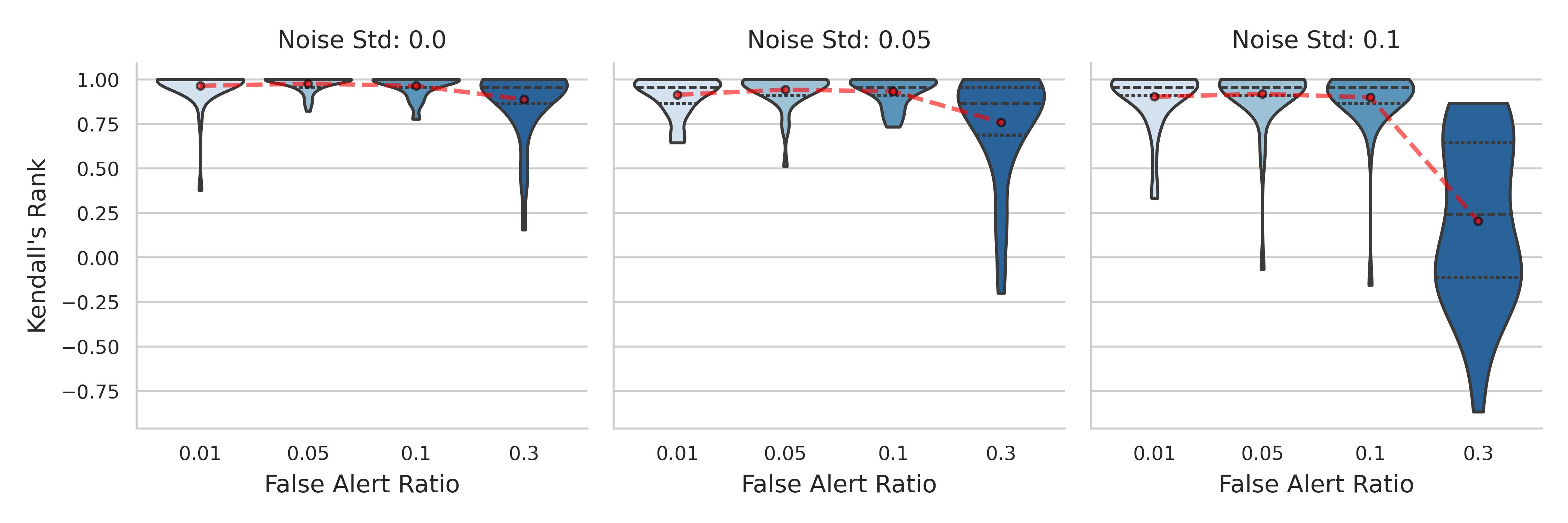}}\\
\subfloat[AUC-ROC\label{fig:robust_aucroc}]{\includegraphics[width=0.5\linewidth]{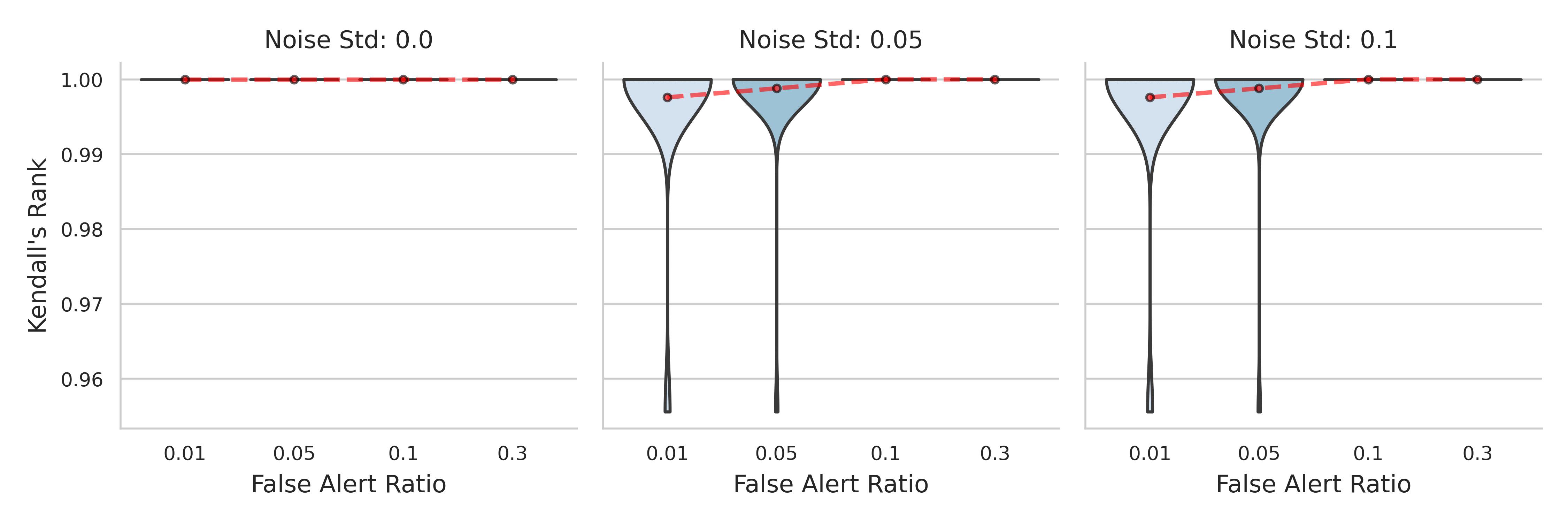}}
\subfloat[VUS-ROC\label{fig:robust_vusroc}]{\includegraphics[width=0.5\linewidth]{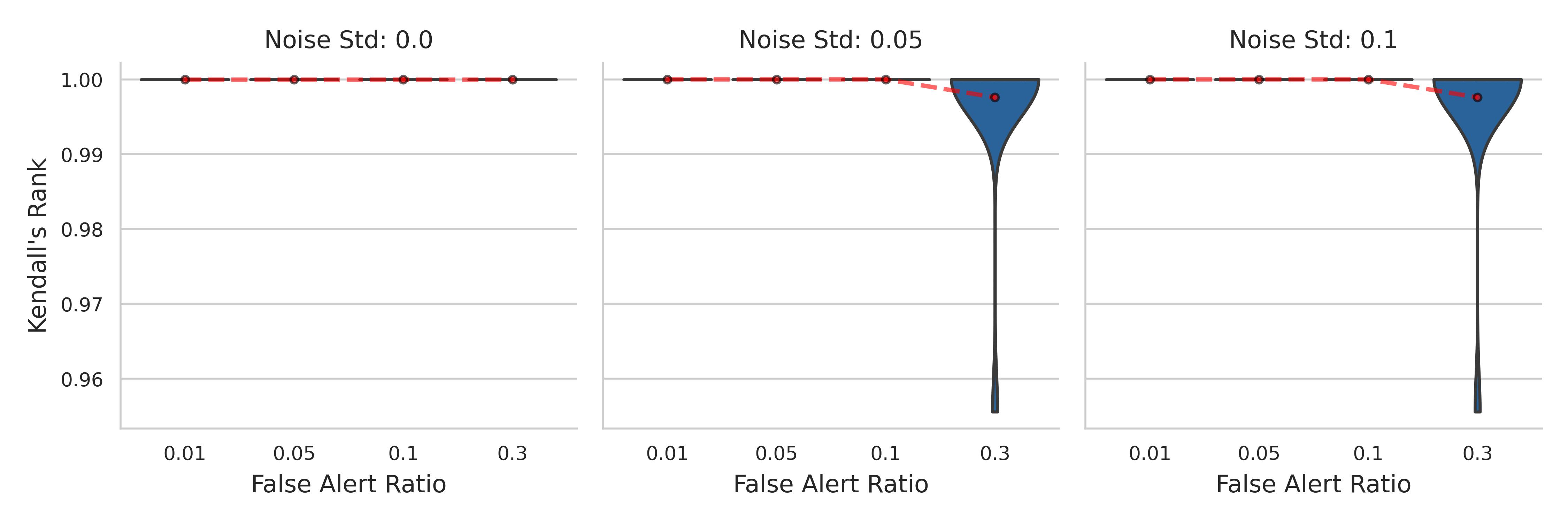}}
\caption{Robustness of different metrics to noise.}
\end{figure*}

Figs. (\ref{fig:robust_cce}-\ref{fig:robust_vusroc}) demonstrate the robustness of different metrics to noise. Under noise-free conditions (i.e., noise std=0), both AUC-ROC and VUS-ROC maintain their ranking capabilities without degradation. However, when noise is present, AUC-ROC exhibits ranking errors at low false positive rates, while VUS-ROC shows ranking errors at high false positive rates. In contrast, Aff-F1 consistently demonstrates ranking inconsistency issues regardless of noise presence, and its error increases progressively with rising false positive rates. This phenomenon occurs because Aff-F1 is based on interval membership, where increased false positives may be interpreted as valid warnings, leading to inflated Aff-F1 scores and consequently introducing experimental bias.

\ifx\includedfrommain\undefined
    \bibliography{main}
    \bibliographystyle{IEEEtran}
    \end{document}
\fi
\fi

\end{document}